\documentclass[pdflatex,sn-mathphys-num]{sn-jnl}

\usepackage{graphicx}%
\usepackage{multirow}%
\usepackage{amsmath,amssymb,amsfonts}%
\usepackage{amsthm}%
\usepackage{mathrsfs}%
\usepackage[title]{appendix}%
\usepackage{xcolor}%
\usepackage{textcomp}%
\usepackage{manyfoot}%
\usepackage{booktabs}%
\usepackage{algorithm}%
\usepackage{algorithmicx}%
\usepackage{algpseudocode}%
\usepackage{listings}%
\usepackage{subcaption}%
\usepackage{adjustbox}
\usepackage{tabularx}
\usepackage{threeparttable}
\usepackage{geometry}
\usepackage{hyperref}
\usepackage{caption}
\usepackage[algo2e]{algorithm2e}

\begin{document}

\title[Article Title]{Brand Visibility in Packaging: A Deep Learning Approach for Logo Detection, Saliency-Map Prediction, and Logo Placement Analysis}

\author[1]{\fnm{Alireza} \sur{Hosseini}}\email{arhosseini77@ut.ac.ir}
\author[1]{\fnm{Kiana} \sur{Hooshanfar}}\email{k.hooshanfar@ut.ac.ir}
\author[2]{\fnm{Pouria} \sur{Omrani}}\email{pouriaomrani@ieee.org}
\author[3]{\fnm{Reza} \sur{Toosi}}\email{rtoosi81@gmail.com}
\author[1]{\fnm{Ramin} \sur{Toosi}}\email{r.toosi@ut.ac.ir}
\author[1]{\fnm{Zahra} \sur{Ebrahimian}}\email{z.ebrahimian@ut.ac.ir}
\author*[1]{\fnm{Mohammad Ali} \sur{Akhaee}}\email{akhaee@ut.ac.ir}

\affil*[1]{\orgdiv{School of Electrical and Computer Engineering}, \orgname{University of Tehran}, \orgaddress{\street{College of Engineering}, \city{Tehran}, \country{Iran}}}

\affil[2]{\orgdiv{Faculty of Electrical Engineering}, \orgname{K. N. Toosi University of Technology}, \orgaddress{\city{Tehran}, \country{Iran}}}

\affil[3]{\orgdiv{Department of Computer Engineering}, \orgname{Faculty of Engineering, Golestan University}, \orgaddress{\city{Gorgan}, \country{Iran}}}


\abstract{In the highly competitive area of product marketing, the visibility of brand logos on packaging plays a crucial role in shaping consumer perception, directly influencing the success of the product. This paper introduces a comprehensive framework to measure the brand logo's attention on a packaging design. The proposed method consists of three steps.
The first step leverages YOLOv8 for precise logo detection across prominent datasets, FoodLogoDet-1500 and LogoDet-3K. The second step involves modeling the user's visual attention with a novel saliency prediction model tailored for the packaging context. The proposed saliency model combines the visual elements with text maps employing a transformers-based architecture to predict user attention maps. In the third step, by integrating logo detection with a saliency map generation, the framework provides a comprehensive brand attention score. The effectiveness of the proposed method is assessed module by module, ensuring a thorough evaluation of each component. Comparing logo detection and saliency map prediction with state-of-the-art models shows the superiority of the proposed methods. To investigate the robustness of the proposed brand attention score, we collected a unique dataset to examine previous psychophysical hypotheses related to brand visibility. the results show that the brand attention score is in line with all previous studies. Also, we introduced seven new hypotheses to check the impact of position, orientation, presence of person, and other visual elements on brand attention.  This research marks a significant stride in the intersection of cognitive psychology, computer vision, and marketing, paving the way for advanced, consumer-centric packaging designs.\footnote{The source code and dataset are available at: \url{https://github.com/Arhosseini77/Brand_Attention}}.}




\maketitle

\section{Introduction}
\label{introduction}

In today's dynamic business world, having a strong brand presence is crucial. The visibility of the brand is incredibly important for keeping up with consumer trends and staying competitive. Consumers often shape their perceptions of brands by considering factors such as visual attractiveness, functionality, and the social significance they convey, predominantly relying on visual cues \citep{BlochPeter1995}. For companies striving to establish and maintain a strong market presence, the packaging of their products, as an interface between the brand and the consumer, significantly influences the purchasing process \citep{Ampuero_Vila2006}.

The visual appeal of packaging, along with the prominent display of design elements, contributes to creating a lasting impression on the consumer and nurturing brand recognition. As consumers navigate the diverse market landscape, a well-designed package captures attention and effectively conveys the brand's values and identity, playing a key role in influencing the purchasing decision \citep{Méndez2011}.

Several studies in marketing and consumer behavior have emphasized the role of effective packaging design in promoting brand recognition \citep{stewart1995packaging}. A well-designed packaging has been shown to significantly enhance brand awareness, purchase intent, and sales \citep{Shukla2022}. These investigations thoroughly explore various aspects of packaging design, conducting a detailed examination of elements such as packaging's shape, texture, and color \citep{Riaz2019, Dong2018, Rebollar2015, PIQUERASFISZMAN2013328, raheem2014impact}. Additionally, they explore the strategic considerations of precise positioning of design elements such as logos, aiming to uncover the subtle interactions between these factors and their impact on consumer perception and brand recognition.

Recognizing the impact of visual elements in packaging, particularly logos, on shaping brand recognition and recall is crucial. This visual aspect influences consumer responses, ultimately playing an important factor in determining the success of a product. Logo, as a fundamental visual element, plays an essential role in packaging design, significantly influencing how consumers perceive and remember a brand \citep{girard2013role}. A visually appealing package not only captures the consumer's attention but also enhances the visibility of the brand logo. On the flip side, weaknesses in design can hinder logo visibility, diminishing its potential impact on consumer awareness \citep{Krishna2017, Otterbring2013}.

Understanding the crucial influence of logo visibility on brand awareness highlights the importance of implementing effective methods to enhance logo visibility. Enhancing logo visibility is linked to understanding its strategic placement on the packaging. The positioning of a logo profoundly impacts its visibility, influencing its interaction with other design elements and resonance with consumers. Thus, it is imperative to focus on optimizing logo placement through strategic positioning on packaging. To this end, implementing advanced machine vision techniques to measure logo visibility becomes crucial to amplifying visibility.
Identifying the logo's position within an image is the initial stage in assessing logo visibility \citep{hou2023deep}. The pursuit of brand visibility does not conclude with knowing the location of the logo; it extends to understanding the attention it commands within the consumer's visual field. This is where saliency prediction \citep{Borji2012} emerges as a pivotal metric. Saliency prediction involves forecasting the perceptual prominence of the logo within the overall visual composition of packaging. Understanding the saliency prediction of the logo enables us to quantify its presence and visual impact, offering a detailed understanding of how much attention the brand attracts on the visual journey of consumers.

The proposed method is positioned to provide a comprehensive framework that includes automated logo detection and a thorough analysis of saliency prediction. This approach is crafted to provide businesses with actionable insights aimed at optimizing logo visibility and creating engaging packaging designs that effectively connect with their target audience. It is composed of three key modules. 
The initial module of our design is the brand logo detection,  leveraging the cutting-edge YOLOv8 architecture. This crucial step helps identify and precisely locate brand logos in visual content. Subsequently, the second module, utilizing a CNN-Transformer-based model generates saliency maps, a crucial element of our methodology. These maps highlight specific regions within the visuals that command the highest visual attention. These insights provide valuable information regarding viewer perception and cognitive responses.
The third and concluding module efficiently integrates the outcomes of both logo detection and saliency map generation. This integration yields a score that quantifies the attention that the brand logo attracts within packaging or advertising visuals.
Furthermore, it is noteworthy to mention that this approach has been validated against existing psychophysical studies related to brand logos in packaging. This validation underscores the  capability of the model to simulate human visual attention on brand logos within packaging and advertising imagery accurately. Consequently, this positions our model as a tool for investigating unexplored experiments regarding brand logos in packaging and advertising contexts. Through this approach, the proposed model provides a comprehensive analysis of brand visual attention, enabling businesses to make informed decisions to enhance their brand presence and impact.
Our main contributions are as follows:

\begin{itemize}
    \item We utilized the YOLOv8 framework to create a logo detection model that outperforms others in accuracy and efficiency.
    
    \item A new saliency prediction model, specifically designed for advertising images and packaging considering text maps, is proposed. This model surpasses state-of-the-art models in saliency prediction.
    
    \item An advanced framework is introduced, designed to measure the level of attention directed towards the brand logo in both packaging and promotional images. 
    
    \item A novel brand attention dataset is introduced to be generated based on a cognitive perspective, exploring 12 different hypotheses.
\end{itemize}

The rest of this work is organized into four sections. Section \ref{Related Works} delves into related work in the field, specifically focusing on optimizing logo placement through eye-tracking, brand logo detection, and saliency map prediction. Section \ref{Description of proposed approaches} outlines the materials, methods, and modeling procedures employed in the research. Section \ref{Experiments and results} is dedicated to discussing the experiments conducted and the results obtained. Finally, section \ref{Conclusion} presents the main conclusions of the work, while proposing future directions and potential enhancements for the introduced architecture.


\section{Related Works}
\label{Related Works}

In this section, we will go through the domain of artificial intelligence (AI) and its applications in the field of marketing, specifically focusing on logo placement design in advertising images and packaging. We will also explore the techniques of brand logo detection and saliency map prediction, discussing their relevance to enhancing brand recognition and optimizing advertising effectiveness.

\subsection{Optimizing Logo Placement with Eye-Tracking}

Neuromarketing, an increasingly influential field of study, uniquely utilizes neuroscience knowledge to directly assess product packaging, eliminating the need to depend on consumers' self-reported preferences \citep{Hubert2008}. By incorporating advanced methodologies like neuroimaging and physiological measurements, neuromarketing employs a more direct and objective approach to assessing consumer responses. This represents a notable shift away from traditional survey-based approaches.
A key methodology in neuromarketing is eye tracking \citep{alvino2021consumer}, providing a detailed examination of visual attention patterns. By studying where and how consumers focus their gaze, researchers gain valuable insights into elements that capture attention and drive perception, uncovering processes beyond conscious awareness \citep{Maynard2018, Gofman2009}.

Specific parameters govern visual behavior, with fixations playing a central role in this context. Fixations, characterized by eye movement, represent moments when the visual system actively acquires information \citep{Pertzov2009}. Numerous studies exploring eye movements, the attention mechanism, and consumer behavior have consistently emphasized the importance of analyzing fixations based on their frequency and duration \citep{Nagel2011}. By understanding the patterns and characteristics of fixations, researchers gain insights into how individuals allocate their visual attention and engage with stimuli. This knowledge proves particularly valuable in fields such as neuromarketing, where assessing consumer responses relies on a detailed understanding of visual attention dynamics.

Employing eye-tracking techniques, previous researches underscore the critical role of packaging design, investigating the influence of specific attributes like color, shape, and labeling on consumer perceptions of the product\citep{Ares2010}.
Strategic positioning of packaging design components is central to practical marketing efforts \citep{Rettie2000}. Inadequate placement may cause crucial design elements to go unnoticed, impacting product evaluation \citep{Krishna2017, Otterbring2013}.
An important study reveals a consumer preference for high-power brands when the brand logo is positioned on the upper side of the packaging, contrasting with diminished appeal when placed on the lower side \citep{Riaz2019, Dong2018}. The effectiveness of capturing participants' attention by placing packaging content at the top is emphasized by Rebollar \emph{et al.}\citep{Rebollar2015}.
Building on existing research, Piqueras-Fiszman \emph{et al.}\citep{PIQUERASFISZMAN2013328} explored the impact of packaging shape and images on consumer attention, with a focus on the logo. Their findings showed that squared-shaped packaging significantly heightened attention toward the logo. Additionally, the study demonstrated the substantial influence of incorporating images on capturing consumer attention. This highlights the complex balance required in packaging design to ensure that attention is not only captured but also sustained, emphasizing the need for strategic placement and thoughtful integration of visual elements to prevent essential components from being marginalized.

\subsection{Brand Logo Detection}

Logo detection, a subfield of object detection, has witnessed substantial advancements over the years. In its initial stages, logo detection heavily relied on manually crafted visual attributes, including the Scale-Invariant Feature Transform (SIFT) and the Histogram of Oriented Gradients (HOG), combined with traditional classification models like Support Vector Machines (SVM)\citep{boia2015elliptical, sahbi2013contextdependent, revaud2012correlation}. However, these approaches faced notable constraints. They were time-consuming because of their region-selective search method using sliding windows. They also struggled to handle different types of logos and were not very efficient at adapting to new situations \citep{hou2023deep}. In recent years, deep learning has emerged as the prevailing paradigm for logo detection. These approaches can be categorized into different strategies, including Region-based Convolutional Neural Network (R-CNN) models and YOLO-based models.
R-CNN models \citep{girshick2014rich}, Fast R-CNN \citep{girshick2015fastrcnn}, and Faster R-CNN \citep{ren2015fasterrcnn} have made noteworthy contributions to the field of logo detection. Hoi \emph{et al.}\citep{hoi2015logonet} introduced the Deep Logo-DRCN scheme, which investigated various techniques within the field of deep region-based convolutional networks (DRCN) for improved logo detection. Similarly, Oliveira \emph{et al.}\citep{oliveira2016automatic} proposed an automatic graphic logo detection system based on Fast R-CNN, known for its robustness under unconstrained imaging conditions. Their approach involved utilizing transfer learning and data augmentation to train a CNN model, enabling multiple detection of potential regions containing objects. Additionally, Li \emph{et al.}\citep{li2017graphic} developed Faster R-CNN for logo detection, incorporating transfer learning, data augmentation, and clustering to optimize hyper-parameters and anchor precision in the Region Proposal Network (RPN), resulting in a significant improvement in detection accuracy.

Feature Pyramid Networks (FPN) are crucial in addressing the multi-scale problem in object detection\citep{lin2017fpn}. FPN notably enhances small object detection without escalating computational demands. Recent works have employed FPN to improve logo detection. Meng \emph{et al.}\citep{meng2021adaptiverepresentation} proposed OSF-Logo, incorporating the Regulated Deformable Convolution (RDC) module in a specific layer of FPN. This integration allows adaptive adjustments of convolution kernel positions, facilitating geometric adaptations to logos. In addition, Jin \emph{et al}\citep{jin2020openbrands} developed Brand Net, utilizing FPN to extract multi-scale features for logo recognition. To enhance small object detection in the context of logo recognition, FPN have also been integrated into Detection Transformers (DETR) \citep{carion2020endtoend}.  Velazquez \emph{et al.} \citep{velazquez2021logodetection} integrated FPN into DETR, enhancing small object detection. Nevertheless, this approach results in an increased computational load during backward propagation. More recently, Hou \emph{et al.}\citep{hou2021foodlogodet1500} proposed the Multi-Scale Feature Decoupling Network (MFDNet) to distinguish between multiple logo categories. MFDNet incorporates a Balanced Feature Pyramid (BFP) for merging multi-scale features and a Feature Offset Module (FOM) with an anchor region proposal network for the optimal selection of logo features.

Driven primarily by the compelling demand for speed and real-time object detection applications, You Only Look Once (YOLO) was developed \citep{redmon2016yolo}. YOLO models, known as single-stage detectors, have played a central role in revolutionizing object detection for their ability to achieve both accuracy and speed. Early versions of YOLO, such as YOLOv2 \citep{redmon2017yolo9000} and YOLOv4 \citep{bochkovskiy2020yolov4}, set new benchmarks in the field. More recent iterations, including YOLOv7 \citep{wang2023yolov7} and YOLOv8 \citep{yolo_ultralytics}, represent the current state-of-the-art in object detection. YOLO models are widely employed, particularly in the domain of logo detection. Palecek \emph{et al.}\citep{palecek2021logodetection} presented Scaled YOLOv4, outperforming traditional two-stage models such as Faster R-CNN in both speed and accuracy. It achieved a relative improvement of up to 46\%, running up to twice as fast. Notably, logo detectors utilizing YOLOv7 and YOLOv8 remain unexplored, presenting an opportunity for potential improvements in balancing accuracy and speed, potentially reaching the state-of-the-art in logo detection.

\subsection{Saliency Map Prediction}
Saliency prediction in computer vision involves the identification and anticipation of the most significant or salient regions within an image or video frame, likely to capture human attention. This process holds practical utility in various applications.
CNNs are commonly used for saliency prediction tasks. Kroner \emph{et al.}\citep{kroner2020contextual} introduced an encoder-decoder framework that incorporates several convolutional layers, each set at various dilation rates, to effectively grasp features on multiple scales. Jia \emph{et al.}\citep{jia2020eml} used deep CNN models to extract more useful visual features for saliency prediction. TempSal \citep{aydemir2023tempsal} enables sequential saliency map generation through a temporal information-based model, astutely exploiting human temporal attention patterns. The incorporation of transfer learning principles amplifies the potential of CNN models in the domain of saliency prediction \citep{kummerer2016deepgaze, linardos2021deepgaze}.
The fusion of RNN with CNN represents a hybrid approach in the field of both image and video saliency prediction, as introduced by Droste \emph{et al.}\citep{droste2020unified}.

Researchers have been inspired by the achievements of attention in natural language processing (NLP) and have started applying these models to computer vision tasks such as saliency prediction. Cao \emph{at al.}\citep{cao2020aggregated} proposed a saliency prediction method named VGG-SSM. Their pipeline consists of three parts: feature extraction, multi-level integration, and a self-attention module. They demonstrated that refining global information from deep layers through a self-attention mechanism, in coordination with fine details in distant portions of a feature map, yields a comprehensive data enhancement process. Additionally, Lou \emph{et al.}\citep{lou2022transalnet} developed a transformer-based method with both DenseNet and ResNet backbones. 

The works mentioned earlier were created for general use, while numerous other works have been suggested specifically for advertising purposes. Lévêque \emph{et al.}\citep{8802989} collected an eye-tracking database of video advertising and evaluated their analysis with state-of-the-art deep learning-based saliency models. Liang \emph{et al.}\citep{liang2021fixation} compiled an eye-tracking dataset comprising 1000 advertising images. Subsequently, they introduced a method that incorporates text features within advertising images, which considers the interaction between text region and pictorial region. Kou \emph{et al.}\citep{10016709} proposed confidence scores fusion for saliency prediction in advertising images, which is helpful to improve the robustness and performance. Another study, conducted by Jiang \emph{et al.}\citep{jiang2022does}, introduces the concept of salient Swin-Transformers. In this work, the researchers initially curated a dataset of e-commerce images for saliency prediction tasks. Subsequently, they proposed a novel multi-task learning framework that demonstrated state-of-the-art performance in e-commerce scenarios.


\section{Proposed Method}
\label{Description of proposed approaches}

\begin{figure}[!t]
    \centering
    \includegraphics[width=1\textwidth]{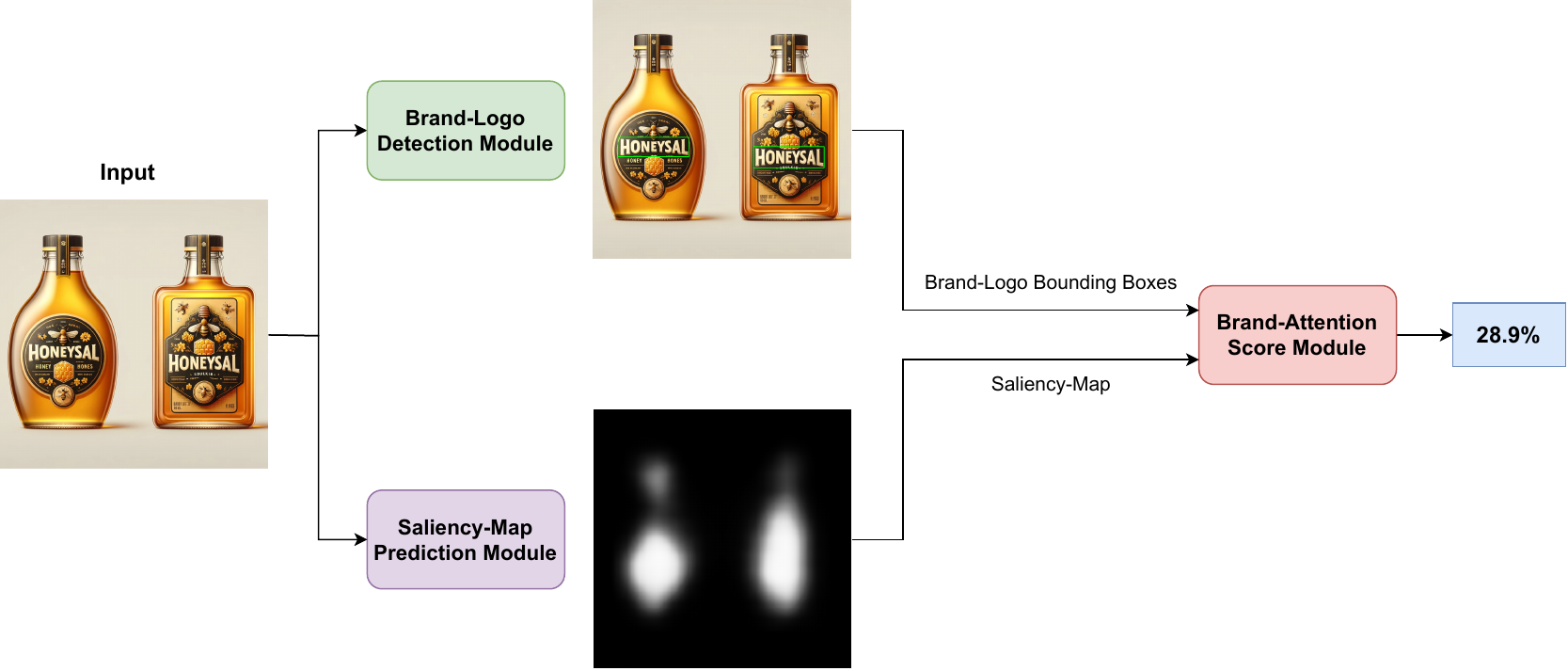}
    \caption{Overview of the proposed Brand-Attention method}
    \label{fig:Brand_Attention_model}
\end{figure}

The primary aim of our research is to design a system for a comprehensive evaluation of the visual prominence of brand logos within the context of packaging or advertising images. To achieve this objective, the proposed methodology encompasses three distinct modules, each designed to address specific aspects of this assessment, as illustrated in \figurename \ref{fig:Brand_Attention_model}. The first module is dedicated to brand logo detection and is supported by the state-of-the-art object detection model, \textit{YOLOv8}. This module identifies and locates brand logos within the imagery, forming the foundational basis for subsequent analysis. Then, the second module focuses on generating saliency maps, a critical aspect of our approach. The saliency maps illuminate the regions within the image that command the highest degree of visual attention, providing valuable insights into viewer perception and cognition. The final module consolidates the outcomes of the brand logo detection and saliency map generation modules. This approach gives a score that measures how much attention the brand logo gets in the packaging or advertising image. This combination of techniques offers valuable insights for businesses aiming to optimize the visual prominence of their brand logos in marketing materials.

\subsection{Brand Logo Detection}

In the initial stage of the proposed method, our focus lies on brand logo detection. For this task, we employ the YOLOv8 model, specifically trained for logo detection purposes. When presented with an input image $I$ with spatial dimensions $H \times W$ and $C$ color channels, our Logo YOLOv8 model processes this image. The output of this model consists of a $1D$ list of bounding boxes, denoted as $B$, where each bounding box ($b$) is represented as a tuple containing the coordinates $(x_{\min}, y_{\min}, x_{\max}, y_{\max})$
\begin{equation}
B = \text{LOGO\_YOLOv8}(I) = [b_1, b_2, \ldots, b_n]
\end{equation}

The number of logo boxes detected in the image is represented by $n$. This detection is the first fundamental step in our brand attention system.

\subsection{Saliency Map Prediction}

\begin{figure}[!t]
    \centering
    \includegraphics[width=1\textwidth]{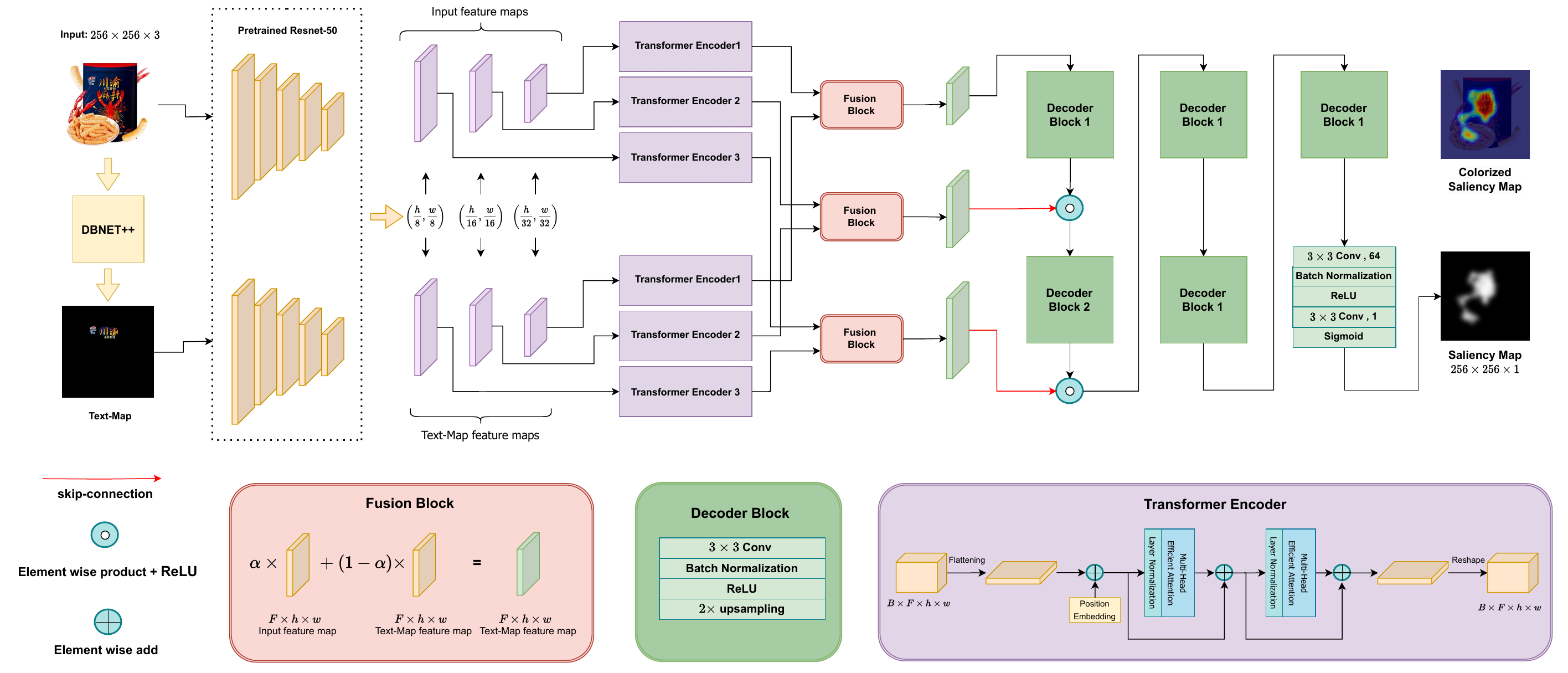}
    \caption{The block diagram of the proposed saliency model.}
    \label{fig:Saliency_Model_Schematic}
\end{figure}

Our primary objective in the second stage is to generate saliency maps for images, with a specific focus on advertising and packaging designs. Previous studies have shown that text is just as important as other visual elements in packaging and advertising. These studies found that text is instrumental in capturing people's attention, and they used eye-tracking data to confirm this \citep{jiang2022does, liang2021fixation}. Recognizing the paramount significance of text in advertising images, we introduce a novel saliency map prediction model tailored to address the unique requirements of both advertisements and packaging images. This model is inspired by the TranSalNet network \citep{lou2022transalnet}, with major improvements made to boost its efficiency and performance. In the proposed model, we initiate the process by detecting text within the image. To achieve this, a pre-trained text detection model is employed \citep{liao2022textdetect}, which outputs a text map. Both the text map and the original image are subsequently processed through a CNN decoder, resulting in multiple feature maps. To efficiently capture and process information from feature maps, we apply transformation through Transformer encoders. This enables the model to consider complex relationships and dependencies within visual content. To ensure a seamless integration of these elements, we introduce a pivotal component of the model: the \textit{Fusion Block}. This block is strategically designed to merge the feature maps derived from both the text map and the original image. By doing so, it enables the simultaneous utilization of visual and text-map features, thereby enhancing the overall interpretative capabilities of the proposed model.
After the fusion block, we use a CNN decoder, which is supported by skip connections coming from the encoder section. This integrated process ensures the restoration of long-range context-enhanced feature maps obtained from the fusion block. These enhanced feature maps serve as the foundation for constructing the final saliency map, capturing the regions of the image that attract the most visual attention.
Figure~\ref{fig:Saliency_Model_Schematic} illustrates the proposed saliency model, providing a visual representation of its architecture and the various components that comprise our refined saliency map prediction system. As depicted, the model comprises five principal components, each of which will be explained in more detail in subsequent sections of this paper.

\subsubsection{Text Detector}

We leverage the cutting-edge DBNet++ network \citep{liao2022textdetect}, which has emerged as a front-runner in the domain of text detection, consistently achieving state-of-the-art accuracy across a spectrum of five scene text detection benchmarks. These benchmarks cover a diverse range of challenges, from handling horizontal and multi-oriented text to curved text, demonstrating the versatility and performance of DBNet++.
The DBNet++ operates on images with spatial dimensions of $H \times W$ and $C$ channels, allowing it to accurately identify text regions within these images. By deploying this innovative network, we can precisely extract and isolate text from non-textual information, ultimately generating text maps.
Given an input image \(I \in \mathbb{R}^{H \times W \times C}\), the DBNet++ detects text regions denoted as $R$. As a consequence, a text map, denoted as \(t_{\text{map}} \in \mathbb{R}^{H \times W \times C},\) is generated as follows:

\begin{equation}
t_{\text{map}}(x, y, c) = 
\begin{cases}
   I(x, y, c) & \text{if } (x, y, c) \in \text{R} \\
   0 & \text{otherwise}
\end{cases}
\end{equation}

\subsubsection{CNN Encoder}
A CNN encoder is employed as our feature extractor. The primary objective of this CNN encoder is to extract essential features from both the image and the text-map while ensuring that the spatial information is distinctly preserved. To achieve this, three sets of convolutional layers is used, each designed to capture features at different spatial scales. Specifically, we extract feature maps with spatial dimensions of \((w/8, h/8)\), \((w/16, h/16)\), and \((w/32, h/32)\). For the image and text-map image feature extraction, the ResNet-50 architecture is utilize \citep{he2016Resnet}.

\subsubsection{Transformer Encoder}
After the initial CNN Encoder stage, which focuses on enhancing long-range and contextual information within our data, we deploy three distinct transformer encoders designed to efficiently capture and process this enriched information. In the pipeline, transformer encoders are integrated to handle the unique characteristics of both original images and text-maps. Specifically, three sets of multi-scale feature maps, denoted as $i_1$, $i_2$, and $i_3$, are derived from the image data. These sets have spatial dimensions of $(w/32, h/32)$, $(w/16, h/16)$, and $(w/8, h/8)$, respectively. Each set is then fed into its respective transformer encoder. To adapt the input size of the transformer encoder and reduce computational complexity, we employ $1 \times 1$ convolution layers (conv1$\times$1) with a stride of one. These convolution layers are applied to the input tensors, including $i_1$, $i_2$, and $i_3$, to decrease their channel dimensions while preserving spatial dimensions. The conv1$\times$1 operation specifically reduces the dimensions of $i_1$, $i_2$, and $i_3$ from 2048, 1024, and 512 to 768, 768, and 512, respectively. This dimension reduction streamlines the data for subsequent processing within the transformer encoder, aligning it with the required input dimensions and optimizing computational efficiency.
Likewise, the textual components of the data, denoted as $t_1$, $t_2$, and $t_3$, undergo dimension reduction through conv1$\times$1 layers employing the same filter size and stride. This process ensures their alignment with the reduced dimensions of the visual components.

To facilitate position awareness and optimize the transformer encoders for effective processing of spatial information within these feature maps, we integrate position embeddings (PE) \citep{dosovitskiy2021POS} into the input before feeding it into the transformer encoders.
Each transformer encoder in the proposed model consists of two identical layers featuring Multi-Head Efficient Attention (MEA) \citep{shen2021efficient} and multi-layer perceptron (MLP) blocks. Notably, the model's design deviates from Transalnet regarding the number of heads and layers in each transformer encoder. Specifically, transformer encoders employ one efficient attention head and a 2-layer MLP. These tailored configurations are designed to meet the specific requirements of our model, ensuring the efficient processing of the enriched feature maps.
Additionally, the MLP block in each transformer encoder consists of two layers with a GELU activation function. Layer normalization (LNorm) and residual connections are applied before and after each block, ensuring stable and effective feature processing.

The introduced methodology distinguishes itself through the adoption of efficient attention, as proposed by Shen \emph{et al.}\citep{shen2021efficient}, diverging from the conventional self-attention mechanism. 
Traditional self-attention is mathematically represented as
\begin{equation}
s(Q, K, V) = \text{softmax}\left(\frac{QK^T}{\sqrt{d_k}}\right)V
\end{equation}

In this formula, \(Q\), \(K\), and \(V\) are the query, key, and value vectors, while \(d_k\) is the embedding dimension. However, this approach is limited by its \(O(N^2)\) computational complexity, which presents major challenges when processing high-resolution images.

\begin{figure}[!ht]
    \centering
    \begin{subfigure}{0.45\textwidth}
        \includegraphics[width=\linewidth]{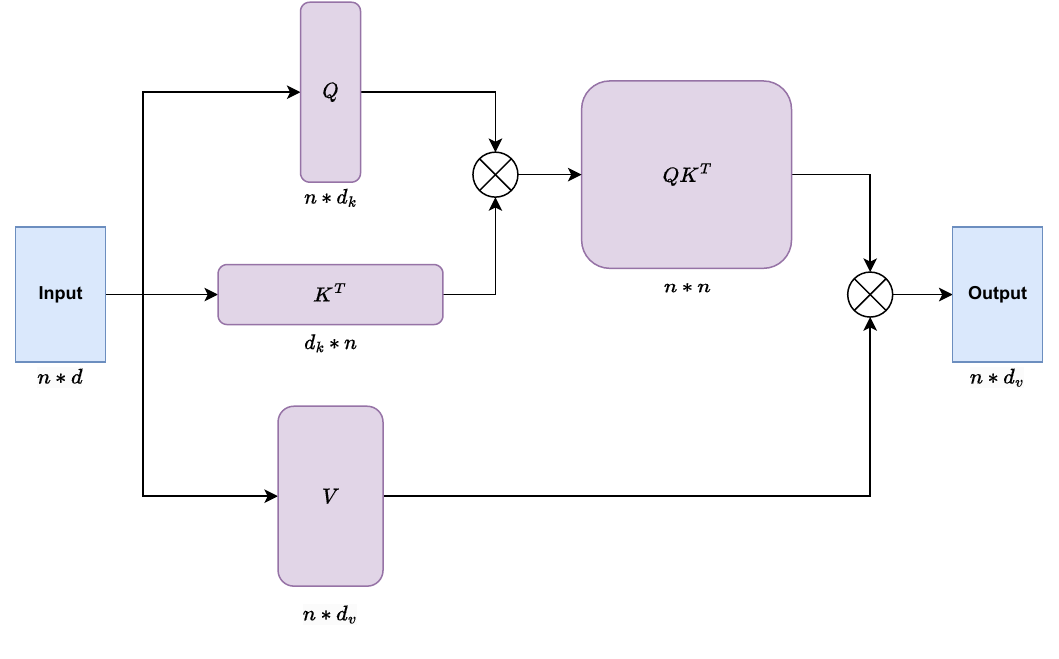}
        \caption{Dot-Product Attention}
        \label{subfig:model1}
    \end{subfigure}
    \hfill
    \begin{subfigure}{0.45\textwidth}
        \includegraphics[width=\linewidth]{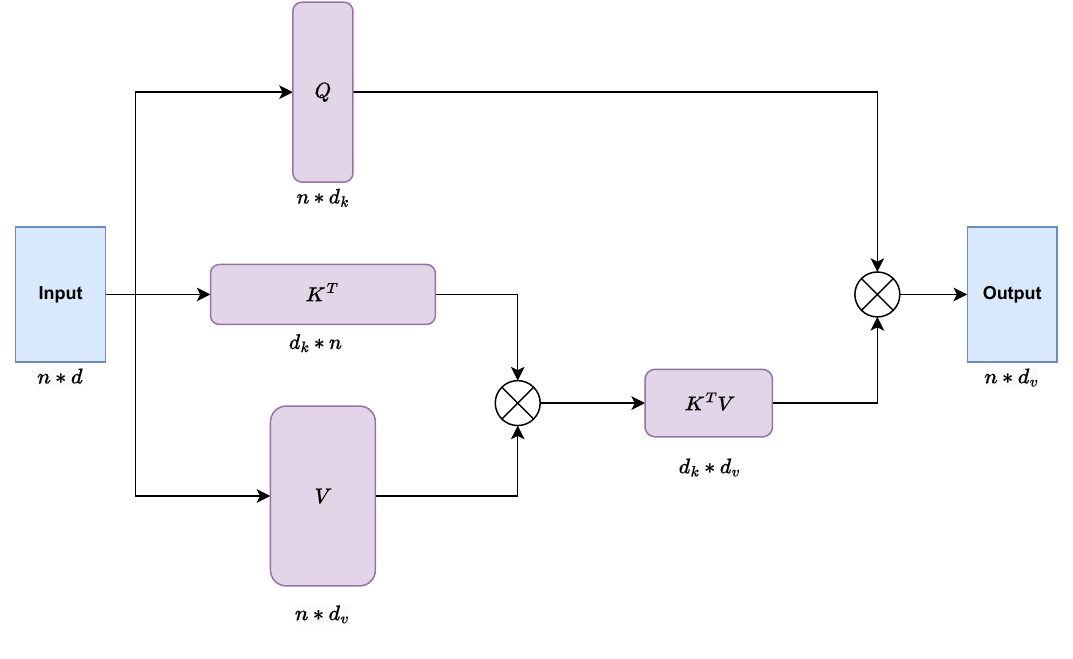}
        \caption{Efficient Attention}
        \label{subfig: model2}
    \end{subfigure}

    \caption{Architecture of dot-product and efficient attention\citep{shen2021efficient}}
    \label{fig: normal and efficient attention}
\end{figure}

Efficient attention, on the other hand, optimizes this process by normalizing the keys and queries before their interaction. Represented as,
\begin{equation}
E(Q, K, V) = \rho_q(Q) (\rho_k(K)^T V)
\end{equation}
where \(\rho_q\) and \(\rho_k\) are normalization functions. This approach addresses the redundancy in the context matrix generation of standard self-attention. It reduces the computational complexity to \(O(d^2n)\), with a memory complexity of \(O(dn + d^2)\), assuming \(d_v = d\) and \(d_k = d/2\). Here, \( d \) represents the embedding dimension.
This model's efficient attention mechanism prioritizes a comprehensive understanding of the input feature, avoiding the computation of pairwise similarities. By treating keys as attention maps \(k_j^T\) and focusing on semantic information rather than positional similarities, it achieves a significant computational efficiency improvement without sacrificing representation richness. The diagram depicting the efficient attention mechanism discussed above is presented in Figure~\ref{fig: normal and efficient attention} \citep{shen2021efficient}.

It can be summarized that for a given sample input $m$ consisting of $t_1$ to $t_3$ (representing textual content) and $i_1$ to $i_3$ (representing image-based features), the transformer encoder process can be mathematically described as follows:
\begin{equation}
z_{0} = \text{conv}_{1\times 1}(m) \oplus \text{PE}
\end{equation}
\begin{equation}
z^{'}_{l} = MEA(\text{LNorm}(z_{l-1}) \oplus z_{l-1})
\end{equation}
\begin{equation}
z_l = \text{MLP}(\text{LNorm}(z^{'}_l) \oplus z^{'}_l)
\end{equation}
where \(z_l\) represents the output feature maps of the \(l\)-th layer in the transformer encoder. The feature maps that go through transformer encoders 1, 2, and 3 are contextually enhanced and are referred to as \(i^{*}_j\) for \(j=1\) to 3 for image and \(t^{*}_j\) for \(j=1\) to 3 for text map image.

\subsubsection{Fusion Block}
After generating enhanced visual features for the image and text map image, it is imperative to merge these features effectively. The fusion process involves assigning weights to the visual and textual modalities. We introduce weighting factors, denoted as $\alpha$, which determine the influence of visual and textual data, respectively.
\begin{equation}
i_{f_j}^{*} = \sigma(\alpha) \cdot i_j^{*} + (1 - \sigma(\alpha)) \cdot t_j^{*}
\end{equation}
In this equation, \(i_f^{*j}\) represents the final feature representation after the fusion process. The selection of the \(\alpha\) parameter is of paramount importance since it governs the equilibrium between the visual and textual modalities. In the proposed model, we treat \(\alpha\) as a learnable parameter, enabling the model to determine the optimal value for this factor. This dynamic approach allows the model to adapt and effectively combine visual and textual information based on the unique demands of the task at hand, thereby enhancing the overall performance and versatility of the model. To ensure that \(\alpha\) remains within the valid range [0, 1] after optimization, a sigmoid function is applied. The sigmoid function, denoted as \(\sigma(\cdot)\), maps real-valued inputs to the interval [0, 1], making it an ideal choice for constraining the \(\alpha\) parameter.

\subsubsection{CNN Decoder}
The CNN decoder plays a key role in integrating and restoring long-range context-enhanced feature maps obtained from the fusion block. Its primary objective is to reconstruct the saliency maps while restoring the original image resolution.
The suggested CNN decoder is designed to facilitate efficient and effective pixel-level classification, enabling the prediction of saliency maps. Within the network, several key operations are performed to enhance the model's performance. After each $3\times 3$ convolution operation (\texttt{Conv3$\times$3}), the batch normalization (\texttt{BNorm}) is applied to promote convergence. Besides, the activation function ReLU is used in all blocks, with Sigmoid employed in the final block.
After initial down-sampling of the input image to a 32-scale by the encoder network, a pivotal process in the CNN Decoder involves a 2-scale up-sampling. This method uses nearest-neighbor interpolation and happens in the first five decoding stages. It creates the saliency map that has the same size as the original input image.

To improve the feature map's long-range and multi-scale context during the decoding process, the up-sampled feature map is fused with the output from the fusion blocks, denoted as \(i_{f_j}^{*}\) for \(j = 1\) to \(3\). This fusion is acquired through the corresponding skip-connection, using an element-wise product operation and ensures that the model benefits from comprehensive contextual information at different scales.

The operations within each CNN decoder block can be represented as follows:
\begin{equation}
O_i = 
\begin{cases}
i_{f_1}^{*}, &  i = 1 \\
\text{ReLU}\left(\text{2X\_Upsample}(O_{i-1}) \cdot i_{f_i}^{*}\right), & i = 2, 3 \\
\text{2X\_Upsample}(O_{i-1}), & i = 4, 5, 6
\end{cases}
\end{equation}
\begin{equation}
O_i^{*} = \text{ReLU}\left(\text{BNorm}(\text{Conv}_{3\times3}(O_i))\right), \quad \text{for } i : 1 \text{ to } 6
\end{equation}
\begin{equation}
S = \text{Sigmoid}(\text{Conv}_{3\times3}(O_6^{*}))
\end{equation}
where \textbf{S} represents the final saliency map predicted by the proposed model.

\subsubsection{Loss Function and Evaluation Metrics}
Drawing inspiration from established conventions in the domain of saliency map prediction models and referencing other saliency prediction frameworks \citep{droste2020unified,che2020gaze, lou2022transalnet}, our model employs a composite loss function. This function combines three metrics: Kullback-Leibler divergence (KL), Linear Correlation Coefficient (CC) and Mean Squared Error (MSE) loss.

Let \(g^{s}\) represent the ground truth of the saliency map, \(g^{f}\) denote the ground truth of the saliency fixation map, and \(S\) denote the network's predicted saliency map. The overarching loss function is defined as:
\begin{equation}
\begin{aligned}
\text{Loss} = & \lambda_1 \cdot \text{KL}(g^{s}, S)
             & + \lambda_2 \cdot \text{CC}(g^{s}, S) 
             & + \lambda_3 \cdot \text{MSELoss}(g^{s}, S)
\end{aligned}
\end{equation}
where each component is elucidated as follows:
\begin{itemize}
    \item \textbf{KL divergence}: A standard measure of dissimilarity between probability distributions, is expressed as:
    \begin{equation}
        \text{KL}(g^{s}, S) = \sum_{i=1}^{n} g^{s}_i \log\left(\epsilon + \frac{S_i}{g^{s}_i + \epsilon}\right)
    \end{equation}
    Here, \(\epsilon\) serves as a regularization constant, set to \(2.2 \times 10^{-16}\).

    \item \textbf{CC}: CC is defined as the ratio of the covariance between \(g^{s}\) and \(S\) to the product of their standard deviations, signifying similarity. The formula is presented as:
    \begin{equation}
        \text{CC}(g^{s}, S) = \frac{\text{cov}(g^{s}, S)}{\sigma(g^{s}) \cdot \sigma(S)}
    \end{equation}
    Here, \(\sigma(.)\) designates the standard deviation, and \(\text{cov}(.)\) stands for the covariance.

\end{itemize}

The objective of this loss function is to minimize the KL and MSELoss while concurrently maximizing the value of CC. This dynamic balance is achieved through the fine-tuning of the coefficients \(\lambda_i\), where \(i\) ranges from 1 to 3.
By employing the Optuna framework \citep{optuna_2019}, we have systematically determined the values for these coefficients to achieve optimal training.
The values of the coefficients are: \(\lambda_1 = 10\), \(\lambda_2 = -3\) and \(\lambda_3 = 5\).
Based on the achieved experiments, these coefficients have been chosen to optimize the model's performance, with a specific focus on reducing KL while concurrently enhancing CC, aligning closely with the intended outcome of the proposed model.

In our comprehensive evaluation framework, we use three additional metrics—Similarity (SIM), Normalized Scan-path Saliency (NSS) and Area under ROC Curve(AUC)—to provide an assessment of the model's performance. While these metrics are not directly embedded within the training loss function, they play an important role in the evaluation phase.
\begin{itemize}
    \item \textbf{SIM}: SIM gauges the linear relationship between the elements of \(g^{s}\) and \(S\), where the minimum value at each position is summed to calculate the coefficient:
    \begin{equation}
        \text{SIM}(g^{s}, S) = \sum_{i=1}^{n} \min(g^{s}_i, S_i)
    \end{equation}

    \item \textbf{NSS}: NSS measures the similarity between the predicted \(S\) and \(g^{f}\) by comparing the fixations with the saliency map values:
    \begin{equation}
    NSS(g^{f}, S) = \frac{1}{\sum_i (g^{f}_i)} \sum_i \left(\frac{S_i - \mu(S)}{\sigma(S)}\right)g^{f}_i
    \end{equation}  
    where, \(\sigma(.)\) designates the standard deviation, and \(\mu(.)\) , \(\text{cov}(.)\) stands for the mean and covariance, respectively.
\end{itemize}

\subsection{Brand-Attention Score}
After localizing brand bounding boxes (B) and generating the saliency maps for both packaging and advertising images, we can quantitatively assess the prominence of the brand within an image. The fundamental concept involves converting the saliency map image into a list of pixel probabilities, ensuring that the cumulative probability sums to 1. Subsequently, we calculate the sum of probabilities associated with pixels contained within the image region.

\begin{algorithm}
\SetAlgoNlRelativeSize{0}
\KwData{B, S}
\KwResult{Brand-Attention Score}
S[S \textless Threshold] = 0 \;

SNorm = S / sum(S) \;

\If{$\mathbf{B}$ is None}{
    \KwRet 0 
}
\Else{
    Brand-Attention Score = 0 \;
    
    \For{b in $\mathbf{B}$}{
        $x_{\text{min}}, y_{\text{min}}, x_{\text{max}}, y_{\text{max}}$ = b \;
        
        \For{y in range($y_{\text{min}}, y_{\text{max}}+1$)}{
            \For{x in range($x_{\text{min}}, x_{\text{max}}+1$)}{
                Brand-Attention Score += SNorm[x, y] \;
            }
        }
    }
    \KwRet Brand-Attention Score \;
}
\caption{Brand-attention score calculation}
\label{algo:BrandAttentionScore}
\end{algorithm}

The pseudo-code for calculating the brand attention score is presented in Algorithm~\ref{algo:BrandAttentionScore}. This pseudo-code outlines the procedure for computing the brand attention score based on the provided saliency map and bounding boxes. It involves removing saliency map values below a threshold, normalizing the remaining values to probabilities, and then calculating the score by summing the normalized values within the specified bounding box regions.
Using the saliency map and this algorithm, we can obtain an attention score for every object or text (not only the brand logo) for which bounding boxes are provided or selected by users.

\section{Experiments and results}\label{Experiments and results}
In this section, we go through the datasets, training setup, and result analysis for both logo detection and saliency prediction. Moreover, the outcomes underscore the enhanced efficacy of the proposed technique compared to leading-edge methods across diverse evaluation metrics. The following part introduces the brand attention module and the proposed dataset. The brand attention module is then validated based on earlier hypotheses, with results thoroughly analyzed using feedback from human participants. The section concludes by proposing and discussing new hypotheses regarding brand visibility in packaging. The computational tasks described in this section were executed using the PyTorch framework on a workstation equipped with an Intel Core i-9 CPU and an NVIDIA GeForce RTX3090 GPU.

\subsection{Logo Detection}
\subsubsection{Datasets}
Over recent periods, the development of specialized datasets for logo detection tasks has garnered considerable focus within the domain of computer vision, providing a valuable resource for various applications \citep{hou2023deep}. To address the challenge of logo detection within the context of diverse packaging products, this research has carefully curated a selection of datasets that align with our specific objectives.

\begin{figure}[ht]
    \centering
    \begin{subfigure}{0.45\textwidth}
        \includegraphics[width=\linewidth]{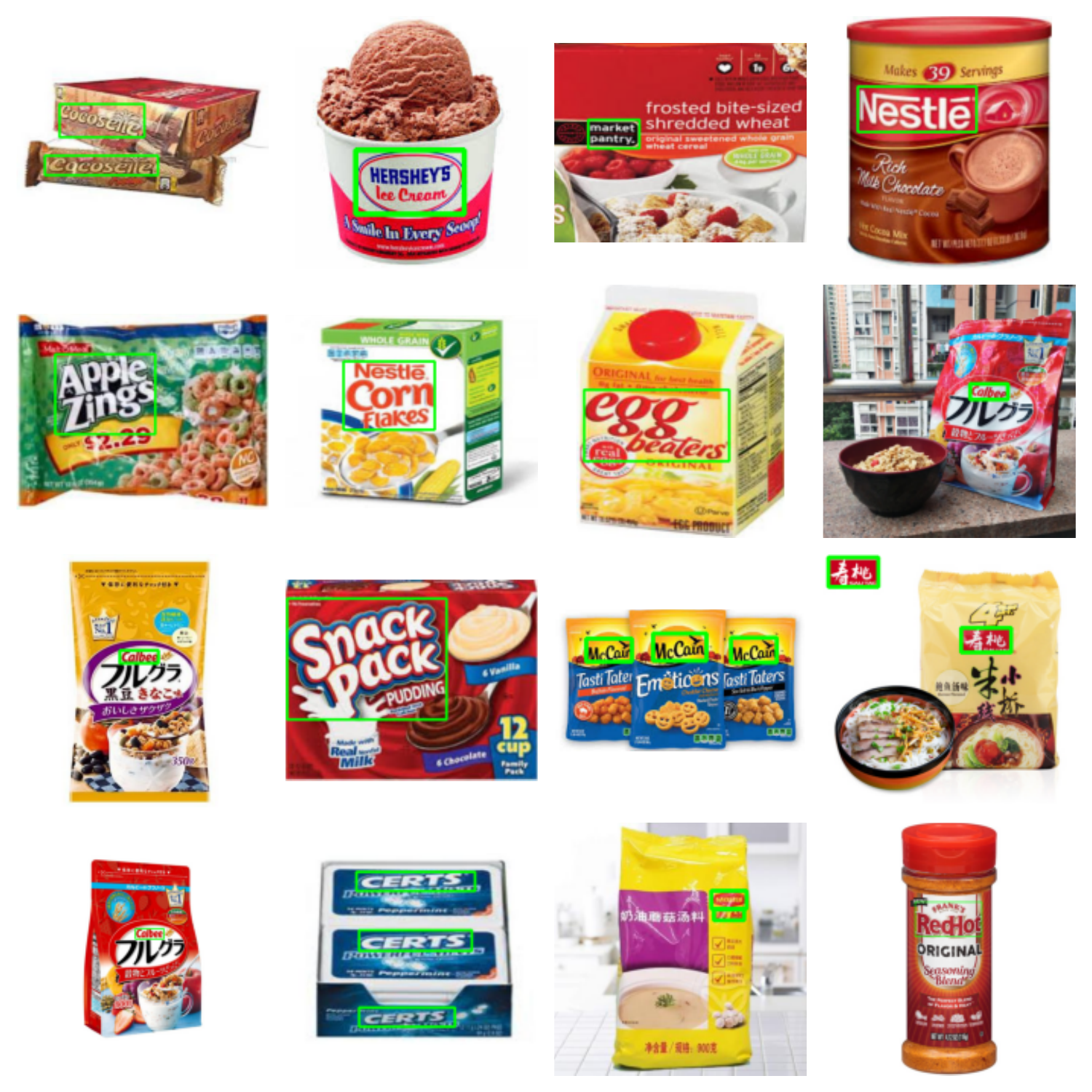}
        \caption{FoodLogoDet-1500}
        \label{subfig:dataset1}
    \end{subfigure}
    \hfill
    \begin{subfigure}{0.45\textwidth}
        \includegraphics[width=\linewidth]{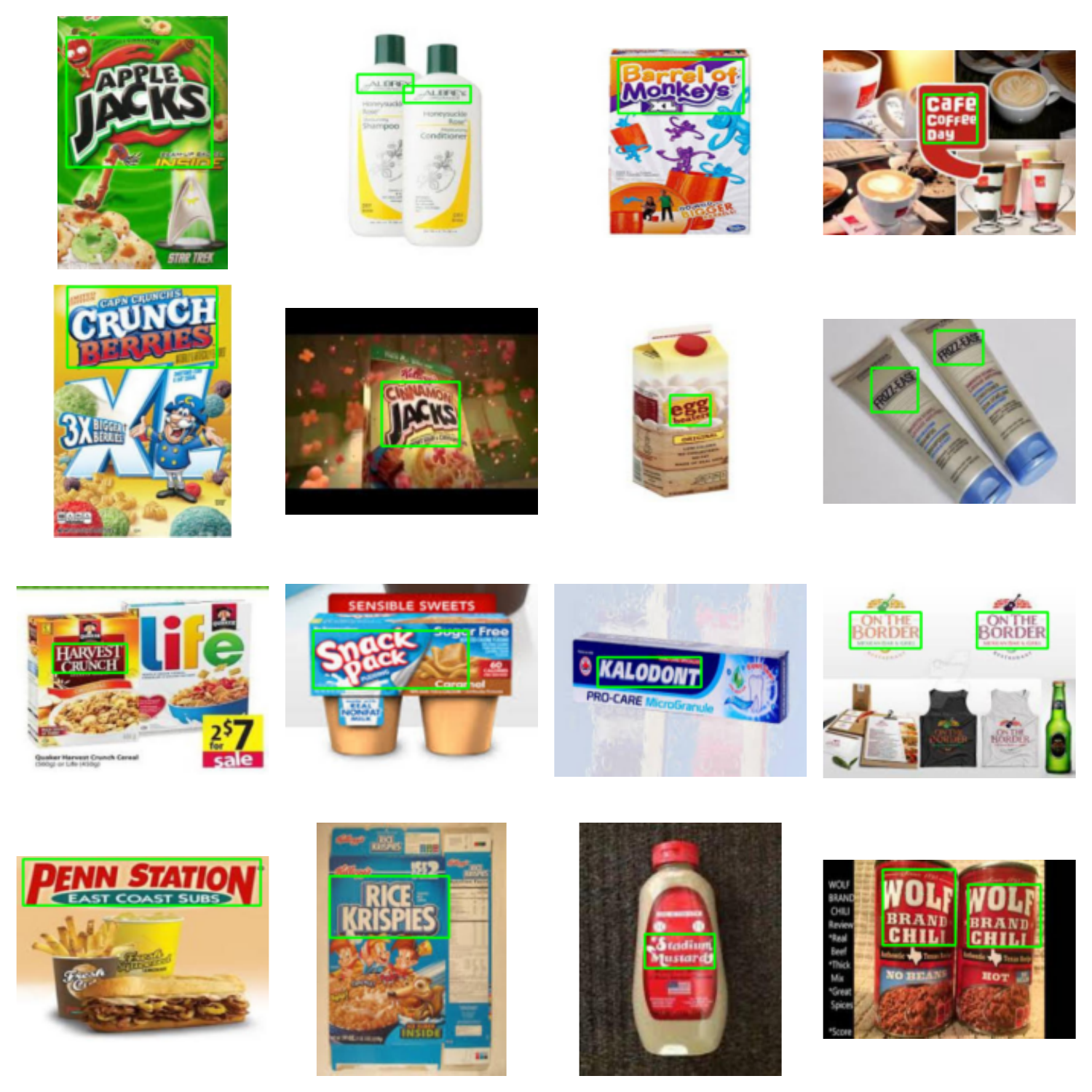}
        \caption{LogoDet-3K}
        \label{subfig:dataset2}
    \end{subfigure}

    \begin{subfigure}{0.45\textwidth}
        \includegraphics[width=\linewidth]{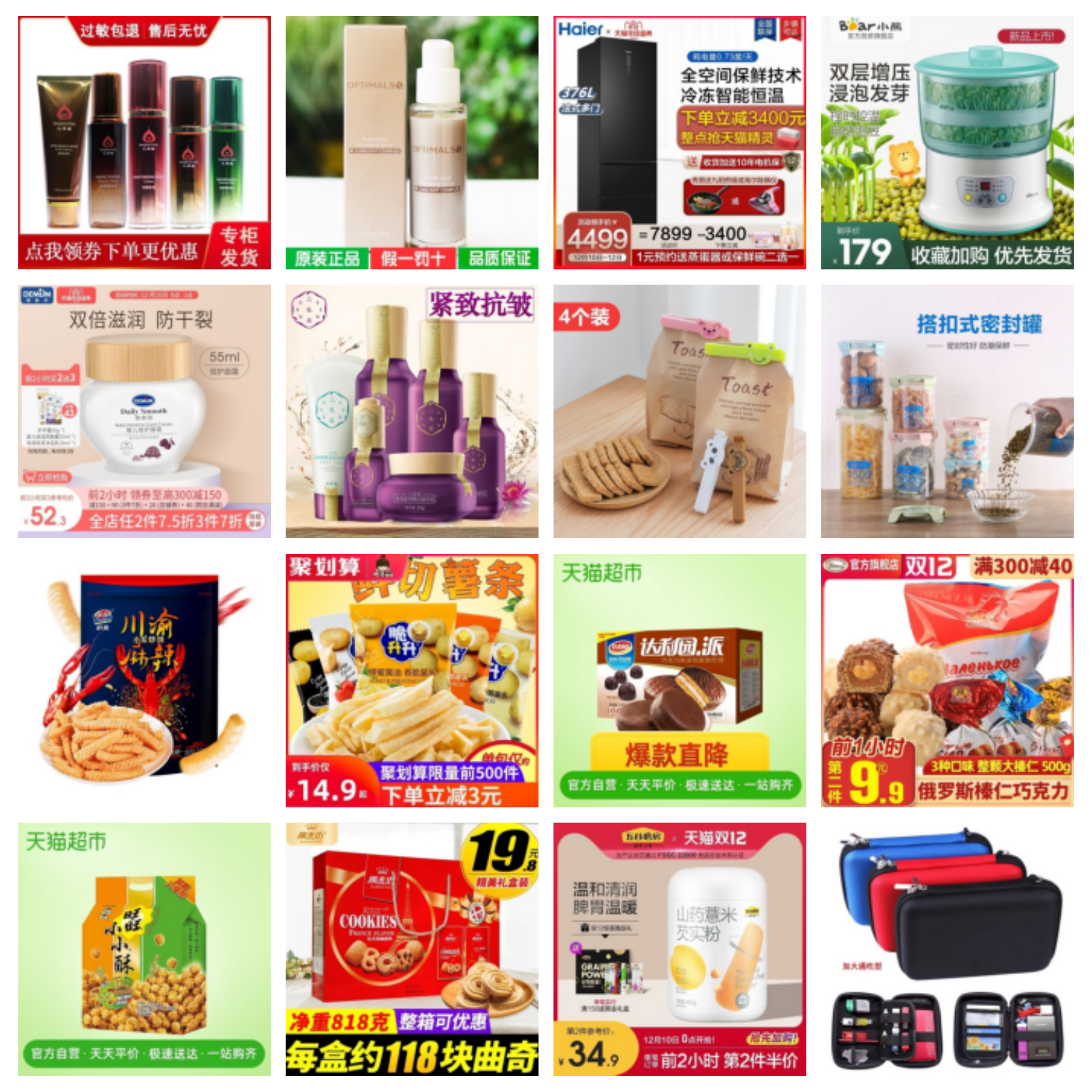}
        \caption{SalECI}
        \label{subfig:dataset3}
    \end{subfigure}

    \caption{Sample images from FoodLogoDet-1500, LogoDet-3K, and SalECI datasets}
    \label{fig:three_datasets}
\end{figure}

Our attention is focused on two logo detection data sets: FoodLogoDet-1500 \citep{hou2021foodlogodet1500}, and LogoDet-3K \citep{wang2022logodet3k}. These datasets were chosen for their unique attributes, making them well-suited for this research and the complexities associated with logo detection in the context of product packaging. A summary of the selected datasets is provided in Table \ref{tab:dataset-summary}. Additionally, to provide a visual perspective, Figure \ref{fig:three_datasets} presents a few samples of the FoodLogoDet-1500 and LogoDet-3K datasets.

\begin{table}[ht]
    \centering
    \caption{Summary of selected logo detection datasets}
    \label{tab:dataset-summary}
    \begin{tabular}{|c|c|c|c|}
    \hline
    \textbf{Dataset} & \textbf{\#Images} & \textbf{\#Objects} & \textbf{\#Logos} \\
    \hline
    FoodLogoDet-1500 & 99,768 & 145,400 & 1,500 \\
    LogoDet-3K & 158,652 & 194,261 & 3,000 \\
    \hline
    \end{tabular}
\end{table}

\subsubsection{Training Setup}
The dataset used for logo detection contains numerous classes, which are not essential for our specific case. Therefore, all classes have been aggregated into one for logo detection. Due to the inability of the proposed model to converge on large-scale datasets, a two-stage fine-tuning process has been implemented. In the first stage, the \textit{small} version of the YOLOv8 model is fine-tuned, initially pre-trained on the COCO dataset, over the FoodLogoDet-1500 dataset. This initial fine-tuning serves as a crucial step to help the model adapt to the characteristics of the data and mitigate convergence issues. The fine-tuning process in this stage is carried out using the Adam optimizer across 100 epochs, with a batch size set to 32, and involves specifying a learning rate of $10^{-2}$ and a momentum of $0.9$.
During the second stage, we continued the fine-tuning process on both the FoodLogoDet-1500 and the larger LogoDet-3k datasets. This approach ensures that the model further adapts to a broader range of data patterns. The second-stage fine-tuning is conducted for 50 epochs with a batch size of 64, using the same hyperparameters as in the first stage. This two-stage fine-tuning strategy has proven effective in addressing the model convergence challenge. The entire process takes approximately 60 hours to complete.

\subsubsection{Result Analysis and Comparison}
We compared the proposed logo detection model with two state-of-the-art methods, including YOLOv7, which was employed as a base line for logo detection, and MFDnet \citep{hou2023deep}. Results are shown in Table \ref{table:logo1} and Table \ref{table:logo2}. As can be observed, YOLOv8 significantly outperforms both YOLOv7 and MFDNet across various metrics, such as mAP50, mAP50-95, precision, and recall, in both stages of evaluation.
\begin{table}[!ht]
\centering
\caption{Metrics on models fine-tunned over Foodlogo-det-1500}
\begin{tabular}{|l|cccc|}
\hline
\textbf{Method} & \textbf{$mAP_{\substack{50}}$} & \textbf{$mAP_{\substack{50-95}}$} & \textbf{Precision} & \textbf{Recall} \\
\hline
\textbf{MFDNet}   & 0.879 & 0.635 & 0.836 & 0.811 \\
\textbf{YOLOv7}     & 0.932 & 0.698 & 0.90 & 0.866 \\
\textbf{YOLOv8}   & \textbf{0.936} & \textbf{0.704} & \textbf{0.904} & \textbf{0.879}  \\
\hline
\end{tabular} \label{table:logo1}
\end{table}

\begin{table}[!ht]
\centering
\caption{Metrics on models which is pretrained on FoodLogo and fine-tunned over FoodLogoDet-1500+LogoDet3k dataset.}
\begin{tabular}{|l|cccc|}
\hline
\textbf{Method} & \textbf{$mAP_{\substack{50}}$} & \textbf{$mAP_{\substack{50-95}}$} & \textbf{Precision} & \textbf{Recall} \\
\hline
\textbf{MFDNet}   & 0.87 & 0.62 & 0.82 & 0.8 \\
\textbf{YOLOv7}     & 0.88 & 0.61 & 0.84 & 0.81 \\
\textbf{YOLOv8}   & \textbf{0.94} & \textbf{0.71} & \textbf{0.91} & \textbf{0.88}  \\
\hline
\end{tabular} \label{table:logo2}
\end{table}

\subsection{Saliency Map Prediction}
\subsubsection{Dataset}
In the domain of saliency map prediction tasks, various general-purpose datasets, including SALICON \citep{jiang2015salicon}, CAT2000 \citep{borji2015cat2000}, MIT1003 \citep{judd2009learning}, and MIT300 \citep{judd2012benchmark} have been established. However, this paper uniquely centers its focus on commercial and advertisement images. To address this specific focus, we leverage the Saliency E-commerce Images (SalECI) dataset introduced by Jiang \emph{et al.} \citep{jiang2022does}.
The SalECI dataset comprises 257,302 fixations obtained through eye-tracking experiments involving 25 subjects. 
The dataset comprises 972 e-commerce images, each paired with corresponding fixation maps and text boundaries. This dataset acts as an important tool for exploring saliency within the realm of commercial and advertising stimuli.

\subsubsection{Training Setup}
The proposed method utilizes 66M parameters, a lower number compared to TranSalNet \citep{lou2022transalnet}. This reduction is achieved through the incorporation of efficient attention mechanisms and a decrease in the number of attention heads and layers. Notably, despite the inclusion of the text detector and fusion block, the overall parameter count remains lower than that of TranSalNet.
During training, the proposed method was trained over the SalECI dataset \citep{jiang2022does} using a step learning rate scheduler with a step size of 4 and a gamma value of 0.1. The initial learning rate was set to $5 \times 10^{-4}$, and weight decay was applied at a rate of $10^{-4}$. The Adam optimizer is used for training.

\subsubsection{Result Analysis and Comparison}
Consistent with the importance of the text map to the proposed saliency prediction method for advertising purposes, a parameter named \(\alpha\) is introduced. The saliency model is initially assessed without a text detector, followed by an evaluation with a fixed \(\alpha\) parameter. Subsequently, the model undergoes another evaluation with the integration of a text detector, maintaining the fixed \(\alpha\) value. Finally, the third evaluation is conducted with a learnable \(\alpha\) parameter, where the initial value of 0.5 dynamically adjusts to 0.659 during the training process. It is noteworthy that the best result is achieved with the \(\alpha\) value of 0.659, which is obtained through the training process.

\begin{figure}[!ht]
    \centering
    \includegraphics[width=1\textwidth]{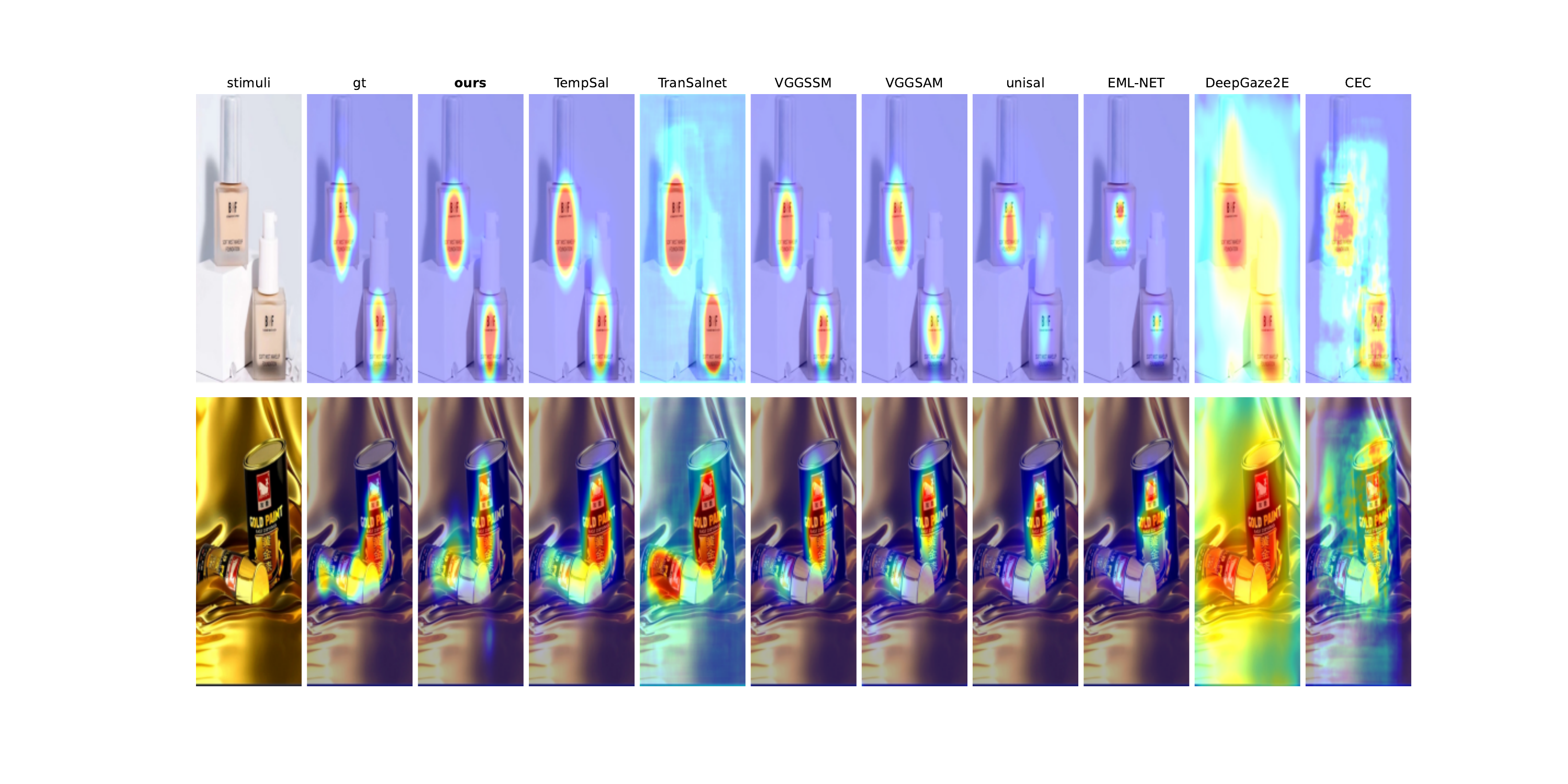}
    \caption{Comparison of the saliency maps of different models over SALECI}
    \label{fig:saliency_compare}
\end{figure}

Table \ref{table:saliency_compare} provides a comparative analysis of the saliency prediction accuracy for ten different state-of-the-art methods using the SalECI dataset. The saliency prediction results of models on the SalECI dataset are visualized in Figure \ref{fig:saliency_compare}.
As seen in Table \ref{table:saliency_compare}, the proposed method emerges as the leading method in this comparison, excelling in correlation coefficient, KL divergence, NSS, and similarity, demonstrating its effectiveness in predicting saliency in the SalECI dataset. These results provide valuable insights into the performance of these methods in the context of saliency prediction.

\begin{table}[!ht]
    \centering
    \caption{Comparing the saliency prediction accuracy for the proposed and nine other state-of-the-art methods over SalECI.}
    \begin{adjustbox}{max width=1\textwidth} 
        \begin{tabular}{|c|ccccc|}
            \hline
            \textbf{Method} & \textbf{CC $\uparrow$} & \textbf{KL $\downarrow$} & \textbf{AUC $\uparrow$} & \textbf{NSS $\uparrow$} & \textbf{SIM $\uparrow$}  \\
            \hline
            Contextual Encoder-Decoder         & 0.459$\pm$0.136 & 1.1346$\pm$0.23 & 0.76$\pm$0.066 & 0.925$\pm$0.268 & 0.373$\pm$0.06  \\
            DeepGazeIIE   & 0.561$\pm$0.124 & 0.995$\pm$0.215 & 0.842$\pm$0.055 & 1.327$\pm$0.318 & 0.399$\pm$0.065  \\
            UNISAL         & 0.6$\pm$0.15 & 0.768$\pm$0.262 & 0.845$\pm$0.056 & 1.574$\pm$0.522 & 0.514$\pm$0.094  \\
            EML-Net        & 0.510$\pm$0.16 & 1.227$\pm$0.903 & 0.807$\pm$0.062 & 1.232$\pm$0.407 & 0.536$\pm$0.103  \\
            VGGSAM        & 0.691$\pm$0.126 & 0.682$\pm$0.259 & 0.815$\pm$0.048 & 1.324$\pm$0.362 & 0.58$\pm$0.091  \\
            Transalnet     & 0.717$\pm$0.061 & 0.873$\pm$0.079 & 0.824$\pm$0.054 & 1.723$\pm$0.203 & 0.534$\pm$0.043  \\
            VGGSSM        & 0.728$\pm$0.121 & 0.599$\pm$0.237 & 0.829$\pm$0.043 & 1.396$\pm$0.359 & 0.611$\pm$0.089  \\
            Temp-SAL       & 0.719$\pm$0.065 & 0.712$\pm$0.126 & 0.813$\pm$0.077 & 1.768$\pm$0.182 & 0.629$\pm$0.048  \\
            SSwin transformer & 0.687$\pm$0.175 & 0.652$\pm$0.478 & 0.868$\pm$0.072 & 1.701$\pm$0.497 & 0.606$\pm$0.101  \\
            ours(without text detection) & 0.741$\pm$0.061 & 0.6958$\pm$0.071 & 0.848$\pm$0.071 & 1.86$\pm$0.234 & 0.635$\pm$0.054  \\
            \textbf{Ours}   & \textbf{0.75$\pm$0.050} & \textbf{0.578$\pm$0.117} & \textbf{0.892$\pm$0.033} & \textbf{1.89$\pm$0.204} & \textbf{0.645$\pm$0.040}  \\
            \hline
        \end{tabular}\label{table:saliency_compare}
    \end{adjustbox}
\end{table}

\subsection{Brand Attention}

In this section, we evaluate the effectiveness of the proposed brand attention module by comparing it with the observations in psychophysical studies. To test the model, we have designed a dataset where each group of images is the same in every way, apart from one particular logo feature it is examining. Wrapping up this section, we introduce some new hypotheses in this field that have not been explored yet.

\subsubsection{Dataset}
While aiming to validate various hypotheses concerning logo placement and packaging design, we have created a dataset comprising 650 images. This collection is a systematically designed platform for testing various ideas connected to packaging design and how people perceive brands.
To ensure a rich and varied base for our study, 95\% of the images in this dataset are sourced from the Internet templates, complemented by those generated by DALL-E, an advanced AI image generation tool. Each image has been carefully modified to align with specific research questions, with alterations ranging from subtle logo repositioning to more substantial design transformations.
Our dataset is organized into 12 hypotheses, each examining different aspects of design and branding. For each hypothesis, we analyze 18 $\pm$ 3 images per hypothesis. In each hypothesis image set, all logo characteristics are fixed, except the one under experiment. This setup provides us with an in-depth insight into the influence of packaging design and logo placement on brand perception.

\subsubsection{Previous Hypothesises Analysis}
We evaluate the effectiveness of our brand attention model by comparing its output to data from human observers who have studied logo attention. This comparison involves aligning the model's predictions with findings from psychophysical studies, ensuring its accuracy in predicting how humans notice logos for real-world applications. The following subsequent items provide a summary of the studies that form the basis of this comparative analysis, showcasing their relevance in the context of brand logo attention:

\begin{figure}[!ht]
    \centering
    \includegraphics[width=1\textwidth]{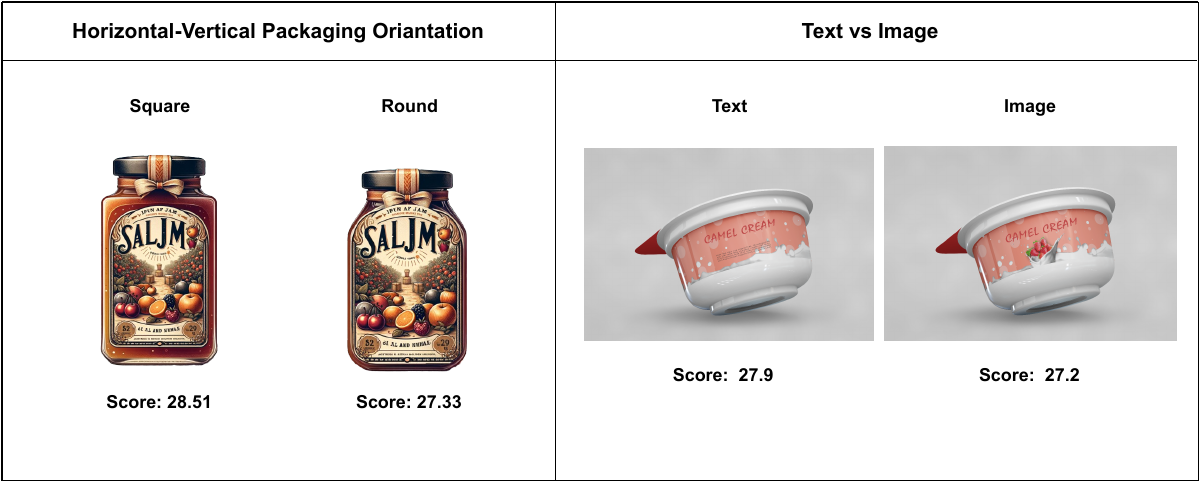}
    \caption{Sample images illustrating the influence of packaging shape (left) and the presence of an image on directing attention to different elements, such as the logo (right).}
    \label{fig:Previous Hypotheses Packagings}
\end{figure}

\begin{figure}[!ht]
    \centering
    \includegraphics[width=1\textwidth]{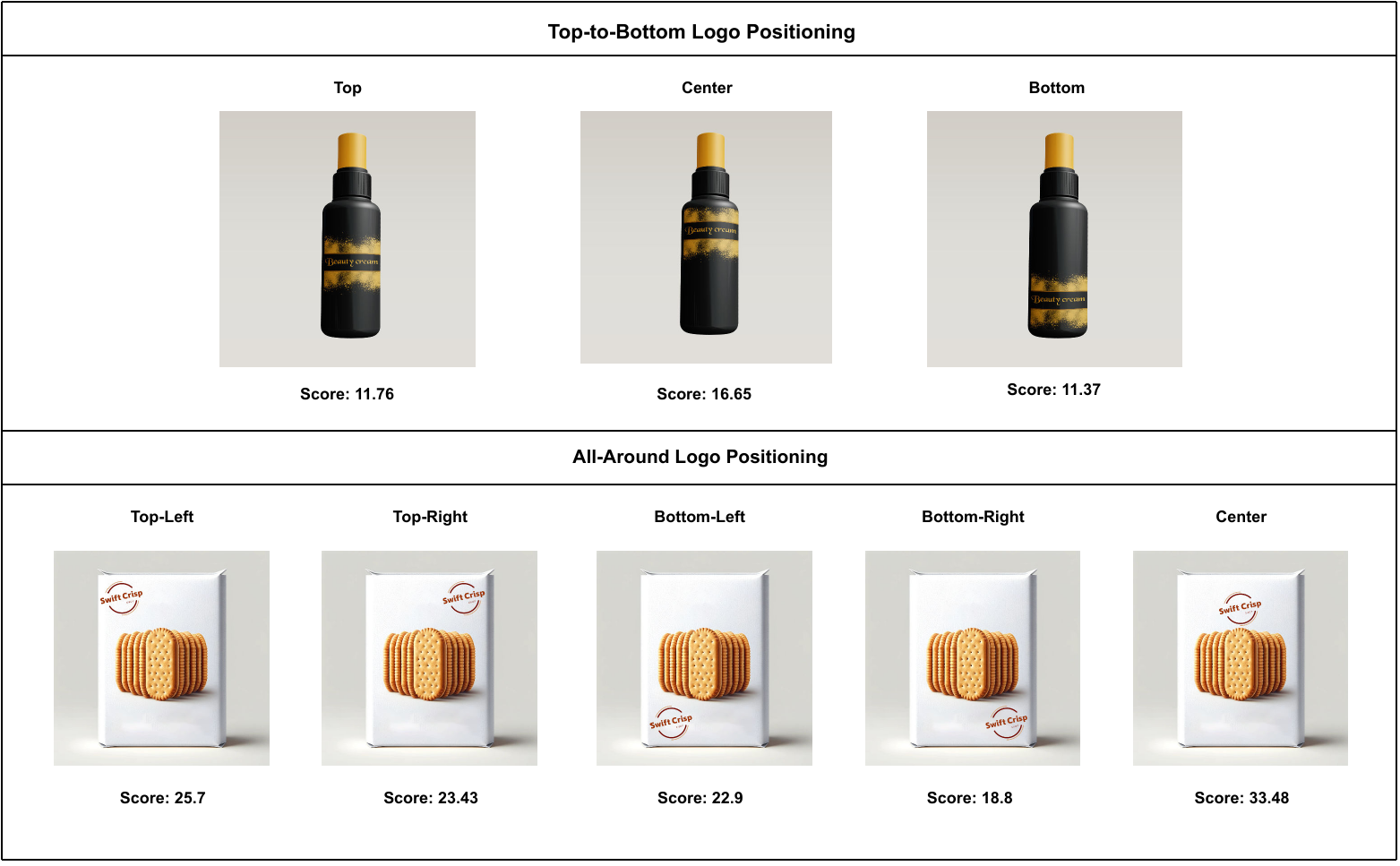}
    \caption{Testing images to demonstrate how logo position impacts brand attention. Top-to-bottom logo positioning (top) and all-around logo positioning (bottom).}
    \label{fig:Positioning of Brand Logos}
\end{figure}

\begin{enumerate}
    \item \textbf{Study 1 }

    Piqueras-Fiszman \emph{et al.}\citep{PIQUERASFISZMAN2013328} examined the influence of packaging shape and the presence of an image on attention to different elements, like the logo. It was found that a squared shape, as opposed to a rounded one, drew more attention to the logo. They have also demonstrated that the photo element on product packaging was highly influential in drawing consumer interest rather than text.
    Backing these ideas, our initial results, as displayed in Table \ref{table:thesis_previous}, suggest that packaging with a squared shape indeed garnered more attention to the logo compared to rounded shapes. Notably, the obtained results align with existing research, underscoring the substantial influence of incorporating visual elements, particularly photos, on product packaging in capturing consumer attention. Figure \ref{fig:Previous Hypotheses Packagings} showcases a sample of images created for testing this study.

    \item \textbf{Study 2}

    The findings from studies proposed by Dong \emph{et al.}\citep{Dong2018} and , Riaz \emph{et al.} \citep{Riaz2019} underscore a noteworthy connection between logo placement and consumer purchase intention. The research reveals a clear preference for high-power brands when logos are positioned at the top, whereas low-power brands garner favor when logos are placed lower. The study explores strategic logo placement using the concept of power metaphors. It suggests that powerful brands should choose top-of-packaging logo placement, aligning with the theory that associates higher placement with increased perceived power. This research holds significance for its implications for rebranding efforts, offering insights into strategic logo placement tailored to brand power dynamics.

    The results of the proposed model, as detailed in Table \ref{table:thesis_previous}, support the observed relationship between logo placement and brand power, reinforcing the practical applicability of these findings in marketing and brand strategy. To further illustrate these concepts, Figure \ref{fig:Positioning of Brand Logos} presents visual examples specifically developed for this thesis.

\end{enumerate}

\subsubsection{Proposed Hypothesises}

Similarly, as outlined in the preceding section, the proposed brand attention method serves as a robust foundation for exploring brand marketing and visual analytics. Beyond the ongoing studies, we have introduced several new hypotheses that investigate aspects not extensively covered in the existing literature. These hypotheses represent unexplored territories as we strive for a comprehensive understanding of brand perception and consumer behavior. This exploration guides future psychophysical tasks, providing a framework for new investigations in the field.

\begin{table}[H]
    \centering
    \caption{comparing the impact of top-to-bottom logo positioning, text vs. image, and square-round packaging orientation hypotheses on brand attention score.}
    \label{table:thesis_previous}
    \begin{tabular}{|l|l|ll|}
    \hline
    \textbf{Hypothesis} & \textbf{Position} & \textbf{Mean} & \textbf{SE} \\ \hline
    
    \multirow{2}{*}{Top-to-Bottom Logo positioning} 
        & Down   & 28.89  & 5.19 \\
        & \textbf{UP}   & 34.05  & 5.64 \\    \hline 
        
    \multirow{2}{*}{Text vs Image}   
        & Image & 31.71  & 4.61 \\
        & \textbf{Text}   & 37.23  & 4.62  \\   \hline
    
    \multirow{2}{*}{Square-Round Packaging Orientation}   
        & Round & 25.82 & 4.86 \\
        & \textbf{Square}   & 27.02  & 4.01  \\   \hline                                      
    
    \end{tabular}
    
\end{table}

\begin{itemize}
    \item \textbf{Positioning of Brand Logos}
    
    Previous studies have mostly looked at where logos are placed, either at the top or bottom of packaging \citep{Dong2018, Riaz2019} and explored the impact of these positions. However, there exists a notable gap in research, specifically addressing central logo placement or exploring the upper-left, bottom-left, upper-right, and lower-right areas. We conducted experiments to investigate the impact of these different placements on brand attention.
    the proposed model predicts that positioning the brand logo at the center of the packaging significantly enhances brand attention when compared to alternative positions, as outlined in Table \ref{table:proposed_thesis}. Additionally, the model proposed that upper placements generally are better than lower ones, with the upper-left being better than the upper-right, likely because people start reading from the top-left \citep{lautenbacher2012still}. Similarly, among the bottom positions, the bottom-left is more effective than the bottom-right.

    \item \textbf{Bold Distinction in Packaging}
    
    Many packaging designs incorporate bold text or objects, yet the impact of these elements on brand logo attention has been under-explored.
    To investigate the impact of emphasized elements other than the brand logo on consumer attention, we conducted experiments and analyzed the findings, as outlined in Table \ref{table:proposed_thesis}. In this experiment, we bolded and emphasized certain non-logo text or objects on the packaging to observe their effect on consumer attention. The proposed model predicts that when elements other than the brand logo are highlighted and emphasized on the packaging, it can lead to a reduction in brand attention. 
        
    \item \textbf{Presence of Person in Packaging}
    
    It is well known that human faces instinctively capture visual attention \citep{cerf2007predicting}. However, the effect of this on brand attention within packaging contexts has not been thoroughly investigated. 
    Exploring the impact of incorporating a person or face on packaging, the introduced model shows a considerable decrease in attention toward the brand. Results in Table \ref{table:proposed_thesis} hint at a diminishing focus on brand logos when human faces are included in the packaging design.

    \item \textbf{Multi Packaging}
    
    The influence of presenting multiple packages of a brand in a single image has not been studied, particularly concerning its impact on brand logo visibility and attention.
    Our model predicts that images containing multiple packages simultaneously of a brand are more effective at absorbing brand attention compared to single-package images as demonstrated in \ref{table:proposed_thesis}. This could lie in the increased likelihood of detecting the brand logo when presented multiple times in diverse packaging formats.
    
    \begin{figure}[!ht]
        \centering
        \includegraphics[width=1\textwidth]{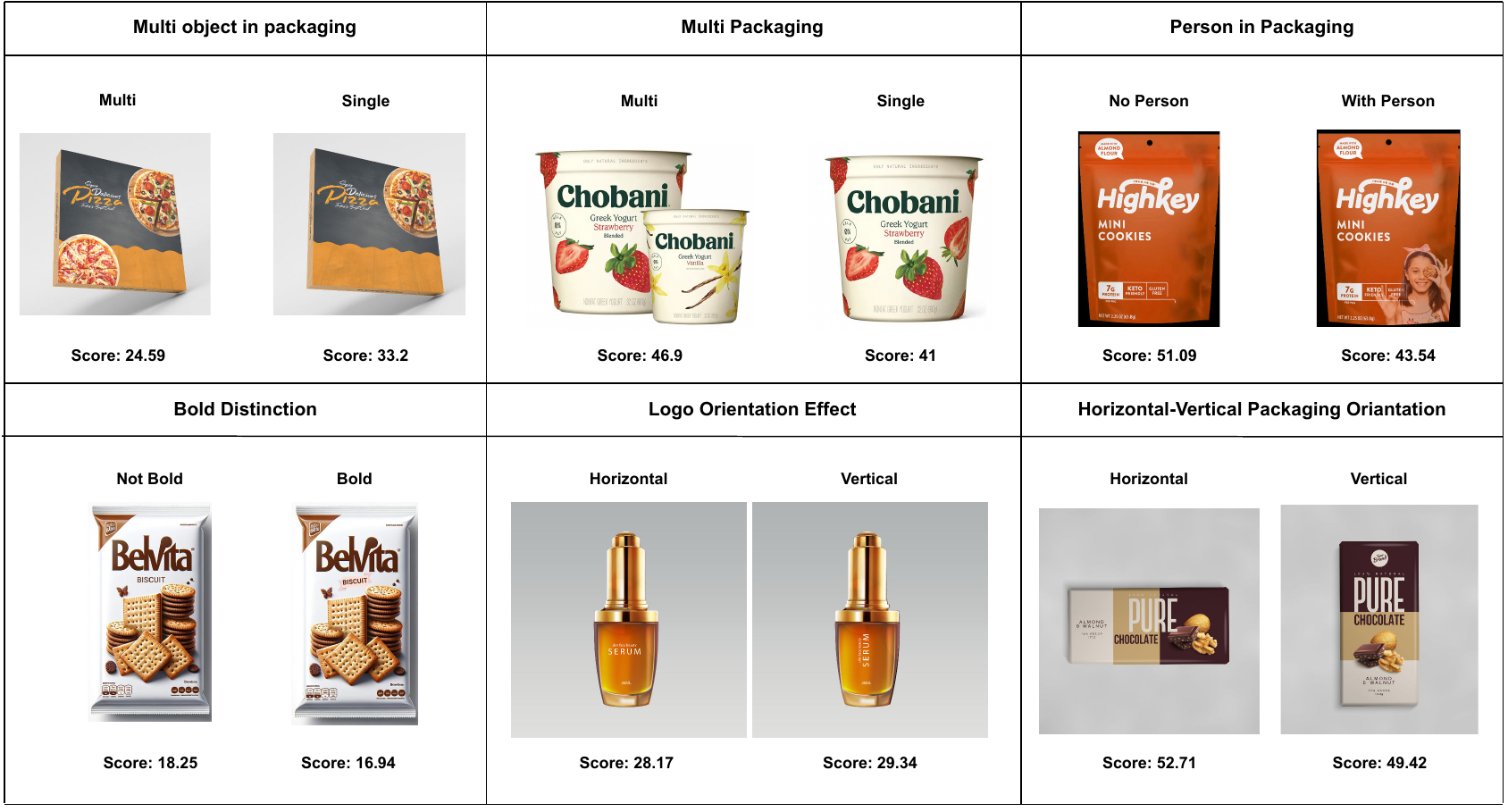}
        \caption{Sample images for assessing the proposed hypotheses: multi objects (top left), multiple packaging (top center), presence of a person (top right), bold distinction (bottom left), horizontal-vertical brand logo orientation (bottom center), and horizontal-vertical packaging orientation (bottom right).}
        \label{fig:Proposed Hypotheses Packagings}
    \end{figure}

    \item \textbf{Multi Objects in Packaging}

    The effect of featuring multiple objects in packaging design (for example, featuring either a single orange or multiple oranges within the packaging), as opposed to just one image, has not been thoroughly investigated in the context of brand logo attention.
    The proposed model predicts that packaging designs featuring multiple objects divert attention from the brand logo, as illustrated in Table \ref{table:proposed_thesis}. The presence of multiple objects in the visual field might cause distraction, leading to a diminished focus on the brand logo.

    This proposition finds support in our experimental results, indicating that simpler packaging designs, featuring only a single object, are more effective in maintaining higher brand logo attention.

    \item \textbf{Horizontal-Vertical Packaging Orientation}

    In terms of the horizontal vs vertical packaging view, the proposed model predicts that horizontally oriented packaging enhances brand logo attention more effectively than vertically oriented packaging. This assumption could be based on the premise that horizontal orientation allows for a larger scale of the brand logo, making it more conspicuous and easily recognizable. The experimental results, as shown in Table \ref{table:proposed_thesis}, indicate a clear preference in brand attention for horizontal packaging designs.

    \item \textbf{Horizontal-Vertical Brand Logo Orientation}
    
    Next, we investigate whether the orientation of the logo itself, horizontally or vertically while keeping other packaging elements constant, would influence the logo attention score. Exploring different orientations, the proposed model suggests that vertical logos grab more attention. This could be from the extra time it takes to read and process vertical text, boosting engagement with the logo \citep{yu2010comparing}. Even when keeping the logo size consistent in both orientations, the proposed model indicates a notable increase in brand attention for vertical logos, as outlined in Table \ref{table:proposed_thesis}.

\begin{table}[H]
    \centering
    \caption{Comparing the impact of top-to-bottom logo positioning, all-around logo positioning, bold distinction, horizontal-vertical brand logo orientation, horizontal-vertical packaging orientation, presence of person, multi-object and multi packaging on brand attention score.}
    \label{table:proposed_thesis}
    \begin{tabular}{|l|l|ll|}
        \hline
        \textbf{Hypothesis} & \textbf{Position} & \textbf{Mean} & \textbf{SE} \\ \hline
        
        \multirow{3}{*}{Top-to-Bottom Logo positioning} 
            & Down   & 28.89  & 5.19 \\
            & UP   & 34.05  & 5.64 \\
            & \textbf{Center}   & 40.02  & 7.06 \\ \hline
        
        \multirow{5}{*}{All-Around Logo Positioning} 
            & Down-Right   & 15.05  & 3.05 \\
            & Down-Left   & 18.8  & 3.41 \\
            & UP-Right   & 16.51  & 3.05 \\
            & UP-Left   & 20.24  & 3.34 \\
            & \textbf{Center}   & 24.92  & 4.12 \\  \hline   
                                                   
        \multirow{2}{*}{Bold Distinction} 
           & Boldness   & 19.98  & 2.27 \\
           & \textbf{Not Bold}   & 21.1  & 2.35 \\ \hline
           
        \multirow{2}{*}{Horizontal-Vertical Brand logo Orientation}  
           & Horizontal & 29.91  & 4.05 \\
           & \textbf{Vertical}   & 34.54  & 4.8  \\ \hline
        
        \multirow{2}{*}{Horizontal-Vertical Packaging Orientation}  
           & Vertical &  27.92 & 4.92 \\
           & \textbf{Horizontal}   &36.92 & 5.59   \\ \hline   
        
        \multirow{2}{*}{Person in Packaging}  
        & With Person & 32.26  & 5.76 \\
        & \textbf{No Person}   & 36  & 6.16  \\  \hline
        
        \multirow{2}{*}{Multi Object in Packaging}  
           & Multi & 32.5  & 5 \\
           & \textbf{One}   & 40.95  & 5.29  \\  \hline
                                                   
        \multirow{2}{*}{Multi Packaging}    
            & Single      & 31.64  & 4.16  \\
            & \textbf{Multi}     & 39.52  & 4.73   \\ \hline
    \end{tabular}
    
\end{table}

\begin{figure}[!ht]
    \centering
    \includegraphics[width=1\textwidth]{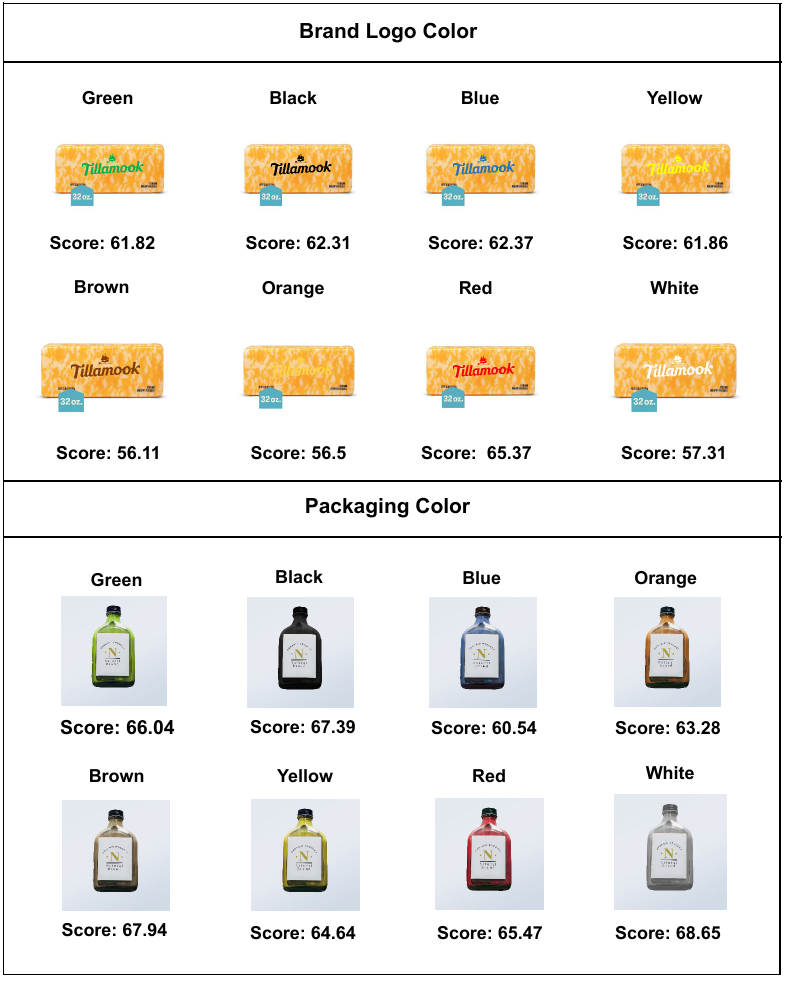}
    \caption{Visualization of the brand-logo color and packaging color influence on brand attention}
    \label{fig:Brand-Logo and Packaging color}
\end{figure}

    \item \textbf{Brand Logo Color}
    
    The influence of different colors of the packaging on attracting attention has been a subject of interest in marketing research \citep{raheem2014impact}. However, comprehensive studies comparing a wide range of colors, particularly in the context of brand logos, are less common.
    This study indicates that logo color influences brand attention, examining colors like white, brown, orange, yellow, green, blue, black, and red. The results, detailed in a table \ref{table:proposed_thesis_color}, show that red is the most effective for drawing attention. This aligns with red's psychological associations with alertness and prominence \citep{singh2006impact}, making it a strategic choice for logos to capture consumer interest.

    \item \textbf{Packaging Color}
    
    Packaging color has a substantial impact on brand visual attention, yet there exists limited research comparing various colors.
    We examined how different packaging colors, including white, brown, orange, yellow, green, blue, black, and red affect the visibility of the brand logo. the proposed model predicts that packaging color greatly affects brand attention, with less intense, warmer, and simpler colors being more effective. The achieved results, as shown in the table \ref{table:proposed_thesis_color}, indicate that white as a packaging color enhances brand attention the most, likely because its neutral and unobtrusive nature allows the logo to be more noticeable.
    
\end{itemize}

\begin{table}[!ht]
    \centering
    \caption{Comparing the impact of packaging color and brand logo color on brand attention score}
    \label{table:proposed_thesis_color}
    \begin{tabular}{|l|l|ll|}
    \hline
    \textbf{Hypothesis} & \textbf{Position} & \textbf{Mean} & \textbf{SE} \\ \hline
    
    \multirow{8}{*}{Packaging Color} 
        & black   & 36.82  & 5.65 \\
        & Brown   & 37.85 & 5.62 \\
        & Orange   & 37.46  & 5.46 \\
        & Yellow   & 37.45  & 5.61 \\
        & Green   & 36.38  & 5.55 \\
        & Blue   & 37.51 & 5.66 \\
        & Red   & 38.23 & 5.73 \\
        & \textbf{White}   & 40.84  & 5.89 \\ \hline
    
    \multirow{8}{*}{Brand Logo Color} 
        & White   & 31.08  & 4.33 \\
        & Brown   & 34.54 & 4.4 \\
        & Orange   & 33.2  & 4.63 \\
        & Yellow   & 32.93  & 4.66 \\
        & Green   & 32.16  & 4.93 \\
        & Blue   & 33.13 & 4.87 \\
        & Black   & 36.56 & 4.7 \\
        & \textbf{Red}   & 37.44  & 4.79 \\ \hline  
                                            
    \end{tabular}
    
\end{table}

Figures \ref{fig:Positioning of Brand Logos} , \ref{fig:Proposed Hypotheses Packagings} and \ref{fig:Brand-Logo and Packaging color} present samples from the brand attention dataset, serving as empirical evidence for the evaluation of our proposed hypotheses.

\section{Conclusion}
\label{Conclusion}
The importance of logos within packaging emerges as an influential visual cue, profoundly shaping consumer perception and promoting brand recognition. This paper introduces a module specifically designed to model human attention to brands in packaging. The module comprises three main components: fine-tuned Yolov8 logo detection, a novel CNN-Transformer-based saliency map prediction model that surpasses existing models in predicting visual attention, and brand attention score. To verify the proposed method, we employed existing psychophysical studies. The proposed method was in line with all previous studies on brand attention, which shows the robustness of the suggested brand attention score.
Next, by utilizing the capabilities of this module, it becomes possible to model human visual attention to brand logos within the packaging, opening new opportunities for testing unexplored hypotheses in this field. Therefore, using the brand attention score, seven new hypothesis, that have never been studied before, are examined. For example, our model suggests that positioning the brand logo at the center or upper left of the packaging increases its visibility. Additionally, the model predicts that employing a red color for the brand logo and a white color for the packaging will enhance the brand's attention score.

The practical utility of the proposed module is especially significant for designers in the fields of advertising and packaging. By quantifying how customers perceive the prominence of a brand, our tool empowers designers to make data-driven decisions regarding logo placement and packaging design. Moreover, the versatility of our module goes beyond brand logos, as designers can use the saliency map and our algorithm to calculate attention scores for any object or text in an image, as long as they designate or select the bounding boxes. This capability enhances the module's applicability, rendering it a valuable tool across various design and marketing applications. Ultimately, this research not only enriches academic discussions on brand visibility but also provides pragmatic tools for enhancing consumer engagement in the dynamic landscape of advertising and packaging.
As the future work, the proposed method can be utilized as a video saliency predictor, considering both spatial and temporal features to enhance its applicability in dynamic contexts. Additionally, the module has the potential to serve as a discrimination network within packaging generative networks, guiding the optimization of logo placement and packaging design. These advancements promise to further revolutionize the field, offering novel insights and tools for effective brand visualization in an ever-evolving digital landscape.

\bibliography{sn-bibliography}


\begin{thebibliography}{71}
\ifx \bisbn   \undefined \def \bisbn  #1{ISBN #1}\fi
\ifx \binits  \undefined \def \binits#1{#1}\fi
\ifx \bauthor  \undefined \def \bauthor#1{#1}\fi
\ifx \batitle  \undefined \def \batitle#1{#1}\fi
\ifx \bjtitle  \undefined \def \bjtitle#1{#1}\fi
\ifx \bvolume  \undefined \def \bvolume#1{\textbf{#1}}\fi
\ifx \byear  \undefined \def \byear#1{#1}\fi
\ifx \bissue  \undefined \def \bissue#1{#1}\fi
\ifx \bfpage  \undefined \def \bfpage#1{#1}\fi
\ifx \blpage  \undefined \def \blpage #1{#1}\fi
\ifx \burl  \undefined \def \burl#1{\textsf{#1}}\fi
\ifx \doiurl  \undefined \def \doiurl#1{\url{https://doi.org/#1}}\fi
\ifx \betal  \undefined \def \betal{\textit{et al.}}\fi
\ifx \binstitute  \undefined \def \binstitute#1{#1}\fi
\ifx \binstitutionaled  \undefined \def \binstitutionaled#1{#1}\fi
\ifx \bctitle  \undefined \def \bctitle#1{#1}\fi
\ifx \beditor  \undefined \def \beditor#1{#1}\fi
\ifx \bpublisher  \undefined \def \bpublisher#1{#1}\fi
\ifx \bbtitle  \undefined \def \bbtitle#1{#1}\fi
\ifx \bedition  \undefined \def \bedition#1{#1}\fi
\ifx \bseriesno  \undefined \def \bseriesno#1{#1}\fi
\ifx \blocation  \undefined \def \blocation#1{#1}\fi
\ifx \bsertitle  \undefined \def \bsertitle#1{#1}\fi
\ifx \bsnm \undefined \def \bsnm#1{#1}\fi
\ifx \bsuffix \undefined \def \bsuffix#1{#1}\fi
\ifx \bparticle \undefined \def \bparticle#1{#1}\fi
\ifx \barticle \undefined \def \barticle#1{#1}\fi
\bibcommenthead
\ifx \bconfdate \undefined \def \bconfdate #1{#1}\fi
\ifx \botherref \undefined \def \botherref #1{#1}\fi
\ifx \url \undefined \def \url#1{\textsf{#1}}\fi
\ifx \bchapter \undefined \def \bchapter#1{#1}\fi
\ifx \bbook \undefined \def \bbook#1{#1}\fi
\ifx \bcomment \undefined \def \bcomment#1{#1}\fi
\ifx \oauthor \undefined \def \oauthor#1{#1}\fi
\ifx \citeauthoryear \undefined \def \citeauthoryear#1{#1}\fi
\ifx \endbibitem  \undefined \def \endbibitem {}\fi
\ifx \bconflocation  \undefined \def \bconflocation#1{#1}\fi
\ifx \arxivurl  \undefined \def \arxivurl#1{\textsf{#1}}\fi
\csname PreBibitemsHook\endcsname

\bibitem[\protect\citeauthoryear{Bloch}{1995}]{BlochPeter1995}
\begin{barticle}
\bauthor{\bsnm{Bloch}, \binits{P.}}:
\batitle{Seeking the ideal form: Product design and consumer response}.
\bjtitle{Journal of Marketing}
\bvolume{59},
\bfpage{16}--\blpage{29}
(\byear{1995})
\doiurl{10.2307/1252116}
\end{barticle}
\endbibitem

\bibitem[\protect\citeauthoryear{Ampuero and Vila}{2006}]{Ampuero_Vila2006}
\begin{barticle}
\bauthor{\bsnm{Ampuero}, \binits{O.}},
\bauthor{\bsnm{Vila}, \binits{N.}}:
\batitle{Consumer perception of product packaging}.
\bjtitle{Journal of Consumer Marketing}
\bvolume{23},
\bfpage{100}--\blpage{112}
(\byear{2006})
\doiurl{10.1108/07363760610655032}
\end{barticle}
\endbibitem

\bibitem[\protect\citeauthoryear{Méndez et~al.}{2011}]{Méndez2011}
\begin{barticle}
\bauthor{\bsnm{Méndez}, \binits{J.}},
\bauthor{\bsnm{Oubiña}, \binits{J.}},
\bauthor{\bsnm{Rubio}, \binits{N.}}:
\batitle{The relative importance of brand-packaging, price and taste in affecting brand preferences}.
\bjtitle{British Food Journal}
\bvolume{113},
\bfpage{1229}--\blpage{1251}
(\byear{2011})
\doiurl{10.1108/00070701111177665}
\end{barticle}
\endbibitem

\bibitem[\protect\citeauthoryear{Stewart}{1995}]{stewart1995packaging}
\begin{bbook}
\bauthor{\bsnm{Stewart}, \binits{B.}}:
\bbtitle{Packaging as an Effective Marketing Tool}.
\bpublisher{CRC Press}, \blocation{???}
(\byear{1995})
\end{bbook}
\endbibitem

\bibitem[\protect\citeauthoryear{Shukla et~al.}{2022}]{Shukla2022}
\begin{barticle}
\bauthor{\bsnm{Shukla}, \binits{P.}},
\bauthor{\bsnm{Singh}, \binits{J.}},
\bauthor{\bsnm{Wang}, \binits{W.}}:
\batitle{The influence of creative packaging design on customer motivation to process and purchase decisions}.
\bjtitle{Journal of Business Research}
\bvolume{147},
\bfpage{338}--\blpage{347}
(\byear{2022})
\doiurl{10.1016/j.jbusres.2022.04.026}
\end{barticle}
\endbibitem

\bibitem[\protect\citeauthoryear{Riaz and Ghafoor}{2019}]{Riaz2019}
\begin{barticle}
\bauthor{\bsnm{Riaz}, \binits{T.}},
\bauthor{\bsnm{Ghafoor}, \binits{M.}}:
\batitle{Strategic logo placement on packaging - using conceptual metaphors of power in packaging – evidence from pakistan}.
\bjtitle{Procedia Computer Science}
\bvolume{158},
\bfpage{582}--\blpage{589}
(\byear{2019})
\doiurl{10.1016/j.procs.2019.09.092}
\end{barticle}
\endbibitem

\bibitem[\protect\citeauthoryear{Dong and Gleim}{2018}]{Dong2018}
\begin{botherref}
\oauthor{\bsnm{Dong}, \binits{R.}},
\oauthor{\bsnm{Gleim}, \binits{M.}}:
High or low: The impact of brand logo location on consumers product perceptions.
Food Quality and Preference
\textbf{69}
(2018)
\doiurl{10.1016/j.foodqual.2018.05.003}
\end{botherref}
\endbibitem

\bibitem[\protect\citeauthoryear{Rebollar et~al.}{2015}]{Rebollar2015}
\begin{barticle}
\bauthor{\bsnm{Rebollar}, \binits{R.}},
\bauthor{\bsnm{Lidón}, \binits{I.}},
\bauthor{\bsnm{Martin~Vallejo}, \binits{F.}},
\bauthor{\bsnm{Puebla}, \binits{M.}}:
\batitle{The identification of viewing patterns of chocolate snack packages using eye-tracking techniques}.
\bjtitle{Food Quality and Preference}
\bvolume{39},
\bfpage{251}--\blpage{258}
(\byear{2015})
\doiurl{10.1016/j.foodqual.2014.08.002}
\end{barticle}
\endbibitem

\bibitem[\protect\citeauthoryear{Piqueras-Fiszman et~al.}{2013}]{PIQUERASFISZMAN2013328}
\begin{barticle}
\bauthor{\bsnm{Piqueras-Fiszman}, \binits{B.}},
\bauthor{\bsnm{Velasco}, \binits{C.}},
\bauthor{\bsnm{Salgado-Montejo}, \binits{A.}},
\bauthor{\bsnm{Spence}, \binits{C.}}:
\batitle{Using combined eye tracking and word association in order to assess novel packaging solutions: A case study involving jam jars}.
\bjtitle{Food Quality and Preference}
\bvolume{28}(\bissue{1}),
\bfpage{328}--\blpage{338}
(\byear{2013})
\doiurl{10.1016/j.foodqual.2012.10.006}
\end{barticle}
\endbibitem

\bibitem[\protect\citeauthoryear{Raheem et~al.}{2014}]{raheem2014impact}
\begin{barticle}
\bauthor{\bsnm{Raheem}, \binits{A.R.}},
\bauthor{\bsnm{Vishnu}, \binits{P.}},
\bauthor{\bsnm{Ahmed}, \binits{A.M.}}:
\batitle{Impact of product packaging on consumer’s buying behavior}.
\bjtitle{European journal of scientific research}
\bvolume{122}(\bissue{2}),
\bfpage{125}--\blpage{134}
(\byear{2014})
\end{barticle}
\endbibitem

\bibitem[\protect\citeauthoryear{Girard et~al.}{2013}]{girard2013role}
\begin{barticle}
\bauthor{\bsnm{Girard}, \binits{T.}},
\bauthor{\bsnm{Anitsal}, \binits{M.M.}},
\bauthor{\bsnm{Anitsal}, \binits{I.}}:
\batitle{The role of logos in building brand awareness and performance: Implications for entrepreneurs}.
\bjtitle{The Entrepreneurial Executive}
\bvolume{18},
\bfpage{7}
(\byear{2013})
\end{barticle}
\endbibitem

\bibitem[\protect\citeauthoryear{Krishna et~al.}{2017}]{Krishna2017}
\begin{barticle}
\bauthor{\bsnm{Krishna}, \binits{A.}},
\bauthor{\bsnm{Cian}, \binits{L.}},
\bauthor{\bsnm{Aydınoğlu}, \binits{N.}}:
\batitle{Sensory aspects of package design}.
\bjtitle{Journal of Retailing}
\bvolume{93},
\bfpage{43}--\blpage{54}
(\byear{2017})
\doiurl{10.1016/j.jretai.2016.12.002}
\end{barticle}
\endbibitem

\bibitem[\protect\citeauthoryear{Otterbring et~al.}{2013}]{Otterbring2013}
\begin{botherref}
\oauthor{\bsnm{Otterbring}, \binits{T.}},
\oauthor{\bsnm{Shams}, \binits{P.}},
\oauthor{\bsnm{Wästlund}, \binits{E.}},
\oauthor{\bsnm{Gustafsson}, \binits{A.}}:
Left isn't always right: Placement of pictorial and textual package elements.
British Food Journal
\textbf{115}
(2013)
\doiurl{10.1108/BFJ-08-2011-0208}
\end{botherref}
\endbibitem

\bibitem[\protect\citeauthoryear{Hou et~al.}{2023}]{hou2023deep}
\begin{barticle}
\bauthor{\bsnm{Hou}, \binits{S.}},
\bauthor{\bsnm{Li}, \binits{J.}},
\bauthor{\bsnm{Min}, \binits{W.}},
\bauthor{\bsnm{Hou}, \binits{Q.}},
\bauthor{\bsnm{Zhao}, \binits{Y.}},
\bauthor{\bsnm{Zheng}, \binits{Y.}},
\bauthor{\bsnm{Jiang}, \binits{S.}}:
\batitle{Deep learning for logo detection: A survey}.
\bjtitle{ACM Transactions on Multimedia Computing, Communications and Applications}
\bvolume{20}(\bissue{3}),
\bfpage{1}--\blpage{23}
(\byear{2023})
\end{barticle}
\endbibitem

\bibitem[\protect\citeauthoryear{Borji and Itti}{2012}]{Borji2012}
\begin{botherref}
\oauthor{\bsnm{Borji}, \binits{A.}},
\oauthor{\bsnm{Itti}, \binits{L.}}:
State-of-the-art in visual attention modeling.
IEEE transactions on pattern analysis and machine intelligence
\textbf{35}
(2012)
\doiurl{10.1109/TPAMI.2012.89}
\end{botherref}
\endbibitem

\bibitem[\protect\citeauthoryear{Hubert et~al.}{2008}]{Hubert2008}
\begin{barticle}
\bauthor{\bsnm{Hubert}, \binits{M.}},
\bauthor{\bsnm{Baecke}, \binits{S.}},
\bauthor{\bsnm{Kenning}, \binits{P.}}:
\batitle{What they see is what they get? an fmri‐study on neural correlates of attractive packaging}.
\bjtitle{Journal of Consumer Behaviour}
\bvolume{7},
\bfpage{342}--\blpage{359}
(\byear{2008})
\doiurl{10.1002/cb.256}
\end{barticle}
\endbibitem

\bibitem[\protect\citeauthoryear{Alvino et~al.}{2021}]{alvino2021consumer}
\begin{barticle}
\bauthor{\bsnm{Alvino}, \binits{L.}},
\bauthor{\bsnm{Constantinides}, \binits{E.}},
\bauthor{\bsnm{Lubbe}, \binits{R.H.}}:
\batitle{Consumer neuroscience: Attentional preferences for wine labeling reflected in the posterior contralateral negativity}.
\bjtitle{Frontiers in psychology}
\bvolume{12},
\bfpage{688713}
(\byear{2021})
\end{barticle}
\endbibitem

\bibitem[\protect\citeauthoryear{Maynard et~al.}{2018}]{Maynard2018}
\begin{botherref}
\oauthor{\bsnm{Maynard}, \binits{O.}},
\oauthor{\bsnm{McClernon}, \binits{F.}},
\oauthor{\bsnm{Oliver}, \binits{J.}},
\oauthor{\bsnm{Munafò}, \binits{M.}}:
Using neuroscience to inform tobacco control policy.
Nicotine \& tobacco research : official journal of the Society for Research on Nicotine and Tobacco
\textbf{21}
(2018)
\doiurl{10.1093/ntr/nty057}
\end{botherref}
\endbibitem

\bibitem[\protect\citeauthoryear{Gofman et~al.}{2009}]{Gofman2009}
\begin{barticle}
\bauthor{\bsnm{Gofman}, \binits{A.}},
\bauthor{\bsnm{Moskowitz}, \binits{H.}},
\bauthor{\bsnm{Fyrbjork}, \binits{J.}},
\bauthor{\bsnm{Moskowitz}, \binits{D.}},
\bauthor{\bsnm{Mets}, \binits{T.}}:
\batitle{Extending rule developing experimentation to perception of food packages with eye tracking}.
\bjtitle{The Open Food Science Journal}
\bvolume{3},
\bfpage{66}--\blpage{78}
(\byear{2009})
\doiurl{10.2174/1874256400903010066}
\end{barticle}
\endbibitem

\bibitem[\protect\citeauthoryear{Pertzov et~al.}{2009}]{Pertzov2009}
\begin{barticle}
\bauthor{\bsnm{Pertzov}, \binits{Y.}},
\bauthor{\bsnm{Avidan}, \binits{G.}},
\bauthor{\bsnm{Zohary}, \binits{E.}}:
\batitle{{Accumulation of visual information across multiple fixations}}.
\bjtitle{Journal of Vision}
\bvolume{9}(\bissue{10}),
\bfpage{2}--\blpage{2}
(\byear{2009})
\doiurl{10.1167/9.10.2}
\end{barticle}
\endbibitem

\bibitem[\protect\citeauthoryear{Nagel et~al.}{2011}]{Nagel2011}
\begin{barticle}
\bauthor{\bsnm{Nagel}, \binits{R.}},
\bauthor{\bsnm{Reutskaja}, \binits{E.}},
\bauthor{\bsnm{Camerer}, \binits{C.}},
\bauthor{\bsnm{Rangel}, \binits{A.}}:
\batitle{Search dynamics in consumer choice under time pressure: An eye-tracking study}.
\bjtitle{American Economic Review}
\bvolume{101},
\bfpage{900}--\blpage{926}
(\byear{2011})
\doiurl{10.1257/aer.101.2.900}
\end{barticle}
\endbibitem

\bibitem[\protect\citeauthoryear{Ares and Deliza}{2010}]{Ares2010}
\begin{barticle}
\bauthor{\bsnm{Ares}, \binits{G.}},
\bauthor{\bsnm{Deliza}, \binits{R.}}:
\batitle{Studying the influence of package shape and colour on consumer expectations of milk desserts using word association and conjoint analysis}.
\bjtitle{Food Quality and Preference}
\bvolume{21},
\bfpage{930}--\blpage{937}
(\byear{2010})
\doiurl{10.1016/j.foodqual.2010.03.006}
\end{barticle}
\endbibitem

\bibitem[\protect\citeauthoryear{Rettie and Brewer}{2000}]{Rettie2000}
\begin{botherref}
\oauthor{\bsnm{Rettie}, \binits{R.}},
\oauthor{\bsnm{Brewer}, \binits{C.}}:
The verbal and visual components of package design.
Journal of Product \& Brand Management
\textbf{9}
(2000)
\doiurl{10.1108/10610420010316339}
\end{botherref}
\endbibitem

\bibitem[\protect\citeauthoryear{Boia et~al.}{2015}]{boia2015elliptical}
\begin{bchapter}
\bauthor{\bsnm{Boia}, \binits{R.}},
\bauthor{\bsnm{Florea}, \binits{C.}},
\bauthor{\bsnm{Florea}, \binits{L.}}:
\bctitle{Elliptical asift agglomeration in class prototype for logo detection}.
In: \bbtitle{Proceedings of the British Machine Vision Conference},
pp. \bfpage{115}--\blpage{111512}
(\byear{2015})
\end{bchapter}
\endbibitem

\bibitem[\protect\citeauthoryear{Sahbi et~al.}{2013}]{sahbi2013contextdependent}
\begin{barticle}
\bauthor{\bsnm{Sahbi}, \binits{H.}},
\bauthor{\bsnm{Ballan}, \binits{L.}},
\bauthor{\bsnm{Serra}, \binits{G.}},
\bauthor{\bsnm{Bimbo}, \binits{A.}}:
\batitle{Context-dependent logo matching and recognition}.
\bjtitle{IEEE Transactions on Image Processing}
\bvolume{22}(\bissue{3}),
\bfpage{1018}--\blpage{1031}
(\byear{2013})
\end{barticle}
\endbibitem

\bibitem[\protect\citeauthoryear{Revaud et~al.}{2012}]{revaud2012correlation}
\begin{bchapter}
\bauthor{\bsnm{Revaud}, \binits{J.}},
\bauthor{\bsnm{Douze}, \binits{M.}},
\bauthor{\bsnm{Schmid}, \binits{C.}}:
\bctitle{Correlation-based burstiness for logo retrieval}.
In: \bbtitle{Proceedings of the 20th ACM International Conference on Multimedia},
pp. \bfpage{965}--\blpage{968}
(\byear{2012})
\end{bchapter}
\endbibitem

\bibitem[\protect\citeauthoryear{Girshick et~al.}{2014}]{girshick2014rich}
\begin{bchapter}
\bauthor{\bsnm{Girshick}, \binits{R.B.}},
\bauthor{\bsnm{Donahue}, \binits{J.}},
\bauthor{\bsnm{Darrell}, \binits{T.}},
\bauthor{\bsnm{Malik}, \binits{J.}}:
\bctitle{Rich feature hierarchies for accurate object detection and semantic segmentation}.
In: \bbtitle{IEEE Conference on Computer Vision and Pattern Recognition},
pp. \bfpage{580}--\blpage{587}
(\byear{2014})
\end{bchapter}
\endbibitem

\bibitem[\protect\citeauthoryear{Girshick}{2015}]{girshick2015fastrcnn}
\begin{bchapter}
\bauthor{\bsnm{Girshick}, \binits{R.B.}}:
\bctitle{Fast r-cnn}.
In: \bbtitle{IEEE International Conference on Computer Vision},
pp. \bfpage{1440}--\blpage{1448}
(\byear{2015})
\end{bchapter}
\endbibitem

\bibitem[\protect\citeauthoryear{Ren et~al.}{2015}]{ren2015fasterrcnn}
\begin{barticle}
\bauthor{\bsnm{Ren}, \binits{S.}},
\bauthor{\bsnm{He}, \binits{K.}},
\bauthor{\bsnm{Girshick}, \binits{R.B.}},
\bauthor{\bsnm{Sun}, \binits{J.}}:
\batitle{Faster r-cnn: Towards real-time object detection with region proposal networks}.
\bjtitle{IEEE Transactions on Pattern Analysis and Machine Intelligence}
\bvolume{39}(\bissue{6}),
\bfpage{1137}--\blpage{1149}
(\byear{2015})
\end{barticle}
\endbibitem

\bibitem[\protect\citeauthoryear{Hoi et~al.}{2015}]{hoi2015logonet}
\begin{botherref}
\oauthor{\bsnm{Hoi}, \binits{S.C.H.}},
\oauthor{\bsnm{Wu}, \binits{X.}},
\oauthor{\bsnm{Liu}, \binits{H.}},
\oauthor{\bsnm{Wu}, \binits{Y.}},
\oauthor{\bsnm{Wang}, \binits{H.}},
\oauthor{\bsnm{Xue}, \binits{H.}},
\oauthor{\bsnm{Wu}, \binits{Q.}}:
Logo-net: Large-scale deep logo detection and brand recognition with deep region-based convolutional networks.
arXiv preprint arXiv:1511.02462
(2015)
\end{botherref}
\endbibitem

\bibitem[\protect\citeauthoryear{Oliveira et~al.}{2016}]{oliveira2016automatic}
\begin{bchapter}
\bauthor{\bsnm{Oliveira}, \binits{G.}},
\bauthor{\bsnm{Frazao}, \binits{X.}},
\bauthor{\bsnm{Pimentel}, \binits{A.}},
\bauthor{\bsnm{Ribeiro}, \binits{B.}}:
\bctitle{Automatic graphic logo detection via fast region-based convolutional networks}.
In: \bbtitle{International Joint Conference on Neural Networks},
pp. \bfpage{985}--\blpage{991}
(\byear{2016})
\end{bchapter}
\endbibitem

\bibitem[\protect\citeauthoryear{Li et~al.}{2017}]{li2017graphic}
\begin{bchapter}
\bauthor{\bsnm{Li}, \binits{Y.}},
\bauthor{\bsnm{Shi}, \binits{Q.}},
\bauthor{\bsnm{Deng}, \binits{J.}},
\bauthor{\bsnm{Su}, \binits{F.}}:
\bctitle{Graphic logo detection with deep region-based convolutional networks}.
In: \bbtitle{IEEE Visual Communications and Image Processing},
pp. \bfpage{1}--\blpage{4}
(\byear{2017})
\end{bchapter}
\endbibitem

\bibitem[\protect\citeauthoryear{Lin et~al.}{2017}]{lin2017fpn}
\begin{bchapter}
\bauthor{\bsnm{Lin}, \binits{T.Y.}},
\bauthor{\bsnm{Dollar}, \binits{P.}},
\bauthor{\bsnm{Girshick}, \binits{R.B.}},
\bauthor{\bsnm{He}, \binits{K.}},
\bauthor{\bsnm{Hariharan}, \binits{B.}},
\bauthor{\bsnm{Belongie}, \binits{S.J.}}:
\bctitle{Feature pyramid networks for object detection}.
In: \bbtitle{IEEE Conference on Computer Vision and Pattern Recognition},
pp. \bfpage{936}--\blpage{944}
(\byear{2017})
\end{bchapter}
\endbibitem

\bibitem[\protect\citeauthoryear{Meng et~al.}{2021}]{meng2021adaptiverepresentation}
\begin{barticle}
\bauthor{\bsnm{Meng}, \binits{Y.}},
\bauthor{\bsnm{Hou}, \binits{S.}},
\bauthor{\bsnm{Wang}, \binits{J.}},
\bauthor{\bsnm{Jia}, \binits{W.}},
\bauthor{\bsnm{Zheng}, \binits{Y.}},
\bauthor{\bsnm{Karim}, \binits{A.}}:
\batitle{An adaptive representation algorithm for multi-scale logo detection}.
\bjtitle{Displays}
\bvolume{70},
\bfpage{102090}
(\byear{2021})
\end{barticle}
\endbibitem

\bibitem[\protect\citeauthoryear{Jin et~al.}{2020}]{jin2020openbrands}
\begin{bchapter}
\bauthor{\bsnm{Jin}, \binits{X.}},
\bauthor{\bsnm{Su}, \binits{W.}},
\bauthor{\bsnm{Zhang}, \binits{R.}},
\bauthor{\bsnm{He}, \binits{Y.}},
\bauthor{\bsnm{Xue}, \binits{H.}}:
\bctitle{The open brands dataset: Unified brand detection and recognition at scale}.
In: \bbtitle{IEEE International Conference on Acoustics, Speech and Signal Processing},
pp. \bfpage{4387}--\blpage{4391}
(\byear{2020})
\end{bchapter}
\endbibitem

\bibitem[\protect\citeauthoryear{Carion et~al.}{2020}]{carion2020endtoend}
\begin{bchapter}
\bauthor{\bsnm{Carion}, \binits{N.}},
\bauthor{\bsnm{Massa}, \binits{F.}},
\bauthor{\bsnm{Synnaeve}, \binits{G.}},
\bauthor{\bsnm{Usunier}, \binits{N.}},
\bauthor{\bsnm{Kirillov}, \binits{A.}},
\bauthor{\bsnm{Zagoruyko}, \binits{S.}}:
\bctitle{End-to-end object detection with transformers}.
In: \bbtitle{European Conference on Computer Vision},
pp. \bfpage{213}--\blpage{229}
(\byear{2020})
\end{bchapter}
\endbibitem

\bibitem[\protect\citeauthoryear{Velazquez et~al.}{2021}]{velazquez2021logodetection}
\begin{barticle}
\bauthor{\bsnm{Velazquez}, \binits{D.A.}},
\bauthor{\bsnm{Gonfaus}, \binits{J.M.}},
\bauthor{\bsnm{Rodr\'{i}guez}, \binits{P.}},
\bauthor{\bsnm{Roca}, \binits{F.X.}},
\bauthor{\bsnm{Ozawa}, \binits{S.}},
\bauthor{\bsnm{Gonzalez}, \binits{J.}}:
\batitle{Logo detection with no priors}.
\bjtitle{IEEE Access}
\bvolume{9},
\bfpage{106}--\blpage{998107011}
(\byear{2021})
\end{barticle}
\endbibitem

\bibitem[\protect\citeauthoryear{Hou et~al.}{2021}]{hou2021foodlogodet1500}
\begin{bchapter}
\bauthor{\bsnm{Hou}, \binits{Q.}},
\bauthor{\bsnm{Min}, \binits{W.}},
\bauthor{\bsnm{Wang}, \binits{J.}},
\bauthor{\bsnm{Hou}, \binits{S.}},
\bauthor{\bsnm{Zheng}, \binits{Y.}},
\bauthor{\bsnm{Jiang}, \binits{S.}}:
\bctitle{Foodlogodet-1500: A dataset for large-scale food logo detection via multi-scale feature decoupling network}.
In: \bbtitle{Proceedings of the 29th ACM International Conference on Multimedia},
pp. \bfpage{4670}--\blpage{4679}
(\byear{2021})
\end{bchapter}
\endbibitem

\bibitem[\protect\citeauthoryear{Redmon et~al.}{2016}]{redmon2016yolo}
\begin{bchapter}
\bauthor{\bsnm{Redmon}, \binits{J.}},
\bauthor{\bsnm{Divvala}, \binits{S.K.}},
\bauthor{\bsnm{Girshick}, \binits{R.B.}},
\bauthor{\bsnm{Farhadi}, \binits{A.}}:
\bctitle{You only look once: Unified, real-time object detection}.
In: \bbtitle{IEEE Conference on Computer Vision and Pattern Recognition},
pp. \bfpage{779}--\blpage{788}
(\byear{2016})
\end{bchapter}
\endbibitem

\bibitem[\protect\citeauthoryear{Redmon and Farhadi}{2017}]{redmon2017yolo9000}
\begin{bchapter}
\bauthor{\bsnm{Redmon}, \binits{J.}},
\bauthor{\bsnm{Farhadi}, \binits{A.}}:
\bctitle{Yolo9000: Better, faster, stronger}.
In: \bbtitle{IEEE Conference on Computer Vision and Pattern Recognition},
pp. \bfpage{6517}--\blpage{6525}
(\byear{2017})
\end{bchapter}
\endbibitem

\bibitem[\protect\citeauthoryear{Bochkovskiy et~al.}{2020}]{bochkovskiy2020yolov4}
\begin{botherref}
\oauthor{\bsnm{Bochkovskiy}, \binits{A.}},
\oauthor{\bsnm{Wang}, \binits{C.Y.}},
\oauthor{\bsnm{Liao}, \binits{H.}}:
Yolov4: Optimal speed and accuracy of object detection.
arXiv preprint arXiv:2004.10934
(2020)
\end{botherref}
\endbibitem

\bibitem[\protect\citeauthoryear{Wang et~al.}{2023}]{wang2023yolov7}
\begin{bchapter}
\bauthor{\bsnm{Wang}, \binits{C.-Y.}},
\bauthor{\bsnm{Bochkovskiy}, \binits{A.}},
\bauthor{\bsnm{Liao}, \binits{H.-Y.M.}}:
\bctitle{Yolov7: Trainable bag-of-freebies sets new state-of-the-art for real-time object detectors}.
In: \bbtitle{Proceedings of the IEEE/CVF Conference on Computer Vision and Pattern Recognition}
(\byear{2023})
\end{bchapter}
\endbibitem

\bibitem[\protect\citeauthoryear{Jocher et~al.}{}]{yolo_ultralytics}
\begin{botherref}
\oauthor{\bsnm{Jocher}, \binits{G.}},
\oauthor{\bsnm{Chaurasia}, \binits{A.}},
\oauthor{\bsnm{Qiu}, \binits{J.}}:
YOLO by Ultralytics.
\url{https://github.com/ultralytics/ultralytics}
\end{botherref}
\endbibitem

\bibitem[\protect\citeauthoryear{Pale{\v{c}}ek and Chaloupka}{2021}]{palecek2021logodetection}
\begin{bchapter}
\bauthor{\bsnm{Pale{\v{c}}ek}, \binits{K.}},
\bauthor{\bsnm{Chaloupka}, \binits{J.}}:
\bctitle{Logo detection and identification in system for audio-visual broadcast transcription}.
In: \bbtitle{2021 44th International Conference on Telecommunications and Signal Processing (TSP)},
pp. \bfpage{357}--\blpage{360}
(\byear{2021})
\end{bchapter}
\endbibitem

\bibitem[\protect\citeauthoryear{Kroner et~al.}{2020}]{kroner2020contextual}
\begin{barticle}
\bauthor{\bsnm{Kroner}, \binits{A.}},
\bauthor{\bsnm{Senden}, \binits{M.}},
\bauthor{\bsnm{Driessens}, \binits{K.}},
\bauthor{\bsnm{Goebel}, \binits{R.}}:
\batitle{Contextual encoder--decoder network for visual saliency prediction}.
\bjtitle{Neural Networks}
\bvolume{129},
\bfpage{261}--\blpage{270}
(\byear{2020})
\end{barticle}
\endbibitem

\bibitem[\protect\citeauthoryear{Jia and Bruce}{2020}]{jia2020eml}
\begin{barticle}
\bauthor{\bsnm{Jia}, \binits{S.}},
\bauthor{\bsnm{Bruce}, \binits{N.D.}}:
\batitle{Eml-net: An expandable multi-layer network for saliency prediction}.
\bjtitle{Image and vision computing}
\bvolume{95},
\bfpage{103887}
(\byear{2020})
\end{barticle}
\endbibitem

\bibitem[\protect\citeauthoryear{Aydemir et~al.}{2023}]{aydemir2023tempsal}
\begin{bchapter}
\bauthor{\bsnm{Aydemir}, \binits{B.}},
\bauthor{\bsnm{Hoffstetter}, \binits{L.}},
\bauthor{\bsnm{Zhang}, \binits{T.}},
\bauthor{\bsnm{Salzmann}, \binits{M.}},
\bauthor{\bsnm{S{\"u}sstrunk}, \binits{S.}}:
\bctitle{Tempsal-uncovering temporal information for deep saliency prediction}.
In: \bbtitle{Proceedings of the IEEE/CVF Conference on Computer Vision and Pattern Recognition},
pp. \bfpage{6461}--\blpage{6470}
(\byear{2023})
\end{bchapter}
\endbibitem

\bibitem[\protect\citeauthoryear{K{\"u}mmerer et~al.}{2016}]{kummerer2016deepgaze}
\begin{botherref}
\oauthor{\bsnm{K{\"u}mmerer}, \binits{M.}},
\oauthor{\bsnm{Wallis}, \binits{T.S.}},
\oauthor{\bsnm{Bethge}, \binits{M.}}:
Deepgaze ii: Reading fixations from deep features trained on object recognition.
arXiv preprint arXiv:1610.01563
(2016)
\end{botherref}
\endbibitem

\bibitem[\protect\citeauthoryear{Linardos et~al.}{2021}]{linardos2021deepgaze}
\begin{bchapter}
\bauthor{\bsnm{Linardos}, \binits{A.}},
\bauthor{\bsnm{K{\"u}mmerer}, \binits{M.}},
\bauthor{\bsnm{Press}, \binits{O.}},
\bauthor{\bsnm{Bethge}, \binits{M.}}:
\bctitle{Deepgaze iie: Calibrated prediction in and out-of-domain for state-of-the-art saliency modeling}.
In: \bbtitle{Proceedings of the IEEE/CVF International Conference on Computer Vision},
pp. \bfpage{12919}--\blpage{12928}
(\byear{2021})
\end{bchapter}
\endbibitem

\bibitem[\protect\citeauthoryear{Droste et~al.}{2020}]{droste2020unified}
\begin{bchapter}
\bauthor{\bsnm{Droste}, \binits{R.}},
\bauthor{\bsnm{Jiao}, \binits{J.}},
\bauthor{\bsnm{Noble}, \binits{J.A.}}:
\bctitle{Unified image and video saliency modeling}.
In: \bbtitle{Computer Vision--ECCV 2020: 16th European Conference, Glasgow, UK, August 23--28, 2020, Proceedings, Part V 16},
pp. \bfpage{419}--\blpage{435}
(\byear{2020}).
\bcomment{Springer}
\end{bchapter}
\endbibitem

\bibitem[\protect\citeauthoryear{Cao et~al.}{2020}]{cao2020aggregated}
\begin{bchapter}
\bauthor{\bsnm{Cao}, \binits{G.}},
\bauthor{\bsnm{Tang}, \binits{Q.}},
\bauthor{\bsnm{Jo}, \binits{K.-h.}}:
\bctitle{Aggregated deep saliency prediction by self-attention network}.
In: \bbtitle{Intelligent Computing Methodologies: 16th International Conference, ICIC 2020, Bari, Italy, October 2--5, 2020, Proceedings, Part III 16},
pp. \bfpage{87}--\blpage{97}
(\byear{2020}).
\bcomment{Springer}
\end{bchapter}
\endbibitem

\bibitem[\protect\citeauthoryear{Lou and et~al.}{2022}]{lou2022transalnet}
\begin{barticle}
\bauthor{\bsnm{Lou}, \binits{J.}},
\bauthor{\bsnm{al.}}:
\batitle{Transalnet: Towards perceptually relevant visual saliency prediction}.
\bjtitle{Neurocomputing}
\bvolume{494},
\bfpage{455}--\blpage{467}
(\byear{2022})
\end{barticle}
\endbibitem

\bibitem[\protect\citeauthoryear{Lévêque and Liu}{2019}]{8802989}
\begin{bchapter}
\bauthor{\bsnm{Lévêque}, \binits{L.}},
\bauthor{\bsnm{Liu}, \binits{H.}}:
\bctitle{An eye-tracking database of video advertising}.
In: \bbtitle{2019 IEEE International Conference on Image Processing (ICIP)},
pp. \bfpage{425}--\blpage{429}
(\byear{2019}).
\doiurl{10.1109/ICIP.2019.8802989}
\end{bchapter}
\endbibitem

\bibitem[\protect\citeauthoryear{Liang et~al.}{2021}]{liang2021fixation}
\begin{barticle}
\bauthor{\bsnm{Liang}, \binits{S.}},
\bauthor{\bsnm{Liu}, \binits{R.}},
\bauthor{\bsnm{Qian}, \binits{J.}}:
\batitle{Fixation prediction for advertising images: Dataset and benchmark}.
\bjtitle{Journal of Visual Communication and Image Representation}
\bvolume{81},
\bfpage{103356}
(\byear{2021})
\end{barticle}
\endbibitem

\bibitem[\protect\citeauthoryear{Kou et~al.}{2023}]{10016709}
\begin{barticle}
\bauthor{\bsnm{Kou}, \binits{Q.}},
\bauthor{\bsnm{Liu}, \binits{R.}},
\bauthor{\bsnm{Lv}, \binits{C.}},
\bauthor{\bsnm{Jiang}, \binits{H.}},
\bauthor{\bsnm{Cheng}, \binits{D.}}:
\batitle{Advertising image saliency prediction method based on score level fusion}.
\bjtitle{IEEE Access}
\bvolume{11},
\bfpage{8455}--\blpage{8466}
(\byear{2023})
\doiurl{10.1109/ACCESS.2023.3236807}
\end{barticle}
\endbibitem

\bibitem[\protect\citeauthoryear{Jiang et~al.}{2022}]{jiang2022does}
\begin{bchapter}
\bauthor{\bsnm{Jiang}, \binits{L.}},
\bauthor{\bsnm{Li}, \binits{Y.}},
\bauthor{\bsnm{Li}, \binits{S.}},
\bauthor{\bsnm{Xu}, \binits{M.}},
\bauthor{\bsnm{Lei}, \binits{S.}},
\bauthor{\bsnm{Guo}, \binits{Y.}},
\bauthor{\bsnm{Huang}, \binits{B.}}:
\bctitle{Does text attract attention on e-commerce images: A novel saliency prediction dataset and method}.
In: \bbtitle{Proceedings of the IEEE/CVF Conference on Computer Vision and Pattern Recognition},
pp. \bfpage{2088}--\blpage{2097}
(\byear{2022})
\end{bchapter}
\endbibitem

\bibitem[\protect\citeauthoryear{Liao and et~al.}{2022}]{liao2022textdetect}
\begin{barticle}
\bauthor{\bsnm{Liao}, \binits{M.}},
\bauthor{\bsnm{al.}}:
\batitle{Real-time scene text detection with differentiable binarization and adaptive scale fusion}.
\bjtitle{IEEE Transactions on Pattern Analysis and Machine Intelligence}
\bvolume{45}(\bissue{1}),
\bfpage{919}--\blpage{931}
(\byear{2022})
\end{barticle}
\endbibitem

\bibitem[\protect\citeauthoryear{He et~al.}{2016}]{he2016Resnet}
\begin{bchapter}
\bauthor{\bsnm{He}, \binits{K.}},
\bauthor{\bsnm{Zhang}, \binits{X.}},
\bauthor{\bsnm{Ren}, \binits{S.}},
\bauthor{\bsnm{Sun}, \binits{J.}}:
\bctitle{Deep residual learning for image recognition}.
In: \bbtitle{2016 IEEE Conference on Computer Vision and Pattern Recognition (CVPR)},
pp. \bfpage{770}--\blpage{778}
(\byear{2016})
\end{bchapter}
\endbibitem

\bibitem[\protect\citeauthoryear{Dosovitskiy et~al.}{2021}]{dosovitskiy2021POS}
\begin{bchapter}
\bauthor{\bsnm{Dosovitskiy}, \binits{A.}},
\bauthor{\bsnm{Beyer}, \binits{L.}},
\bauthor{\bsnm{Kolesnikov}, \binits{A.}},
\bauthor{\bsnm{al.}}:
\bctitle{An image is worth 16x16 words: Transformers for image recognition at scale}.
In: \bbtitle{2021 International Conference on Learning Representations (ICLR)}
(\byear{2021})
\end{bchapter}
\endbibitem

\bibitem[\protect\citeauthoryear{Shen and et~al.}{2021}]{shen2021efficient}
\begin{bchapter}
\bauthor{\bsnm{Shen}, \binits{Z.}},
\bauthor{\bsnm{al.}}:
\bctitle{Efficient attention: Attention with linear complexities}.
In: \bbtitle{Proceedings of the IEEE/CVF Winter Conference on Applications of Computer Vision}
(\byear{2021})
\end{bchapter}
\endbibitem

\bibitem[\protect\citeauthoryear{Che et~al.}{2020}]{che2020gaze}
\begin{barticle}
\bauthor{\bsnm{Che}, \binits{Z.}},
\bauthor{\bsnm{Borji}, \binits{A.}},
\bauthor{\bsnm{Zhai}, \binits{G.}},
\bauthor{\bsnm{Min}, \binits{X.}},
\bauthor{\bsnm{Guo}, \binits{G.}},
\bauthor{\bsnm{Callet}, \binits{P.L.}}:
\batitle{Why is gaze influenced by image transformations? dataset and model}.
\bjtitle{IEEE Transactions on Image Processing}
\bvolume{29},
\bfpage{2287}--\blpage{2300}
(\byear{2020})
\end{barticle}
\endbibitem

\bibitem[\protect\citeauthoryear{Akiba et~al.}{2019}]{optuna_2019}
\begin{bchapter}
\bauthor{\bsnm{Akiba}, \binits{T.}},
\bauthor{\bsnm{Sano}, \binits{S.}},
\bauthor{\bsnm{Yanase}, \binits{T.}},
\bauthor{\bsnm{Ohta}, \binits{T.}},
\bauthor{\bsnm{Koyama}, \binits{M.}}:
\bctitle{Optuna: A next-generation hyperparameter optimization framework}.
In: \bbtitle{Proceedings of the 25th {ACM} {SIGKDD} International Conference on Knowledge Discovery and Data Mining}
(\byear{2019})
\end{bchapter}
\endbibitem

\bibitem[\protect\citeauthoryear{Wang et~al.}{2022}]{wang2022logodet3k}
\begin{barticle}
\bauthor{\bsnm{Wang}, \binits{J.}},
\bauthor{\bsnm{Min}, \binits{W.}},
\bauthor{\bsnm{Hou}, \binits{S.}},
\bauthor{\bsnm{Ma}, \binits{S.}},
\bauthor{\bsnm{Zheng}, \binits{Y.}},
\bauthor{\bsnm{Jiang}, \binits{S.}}:
\batitle{Logodet3k: A large-scale image dataset for logo detection}.
\bjtitle{ACM Transactions on Multimedia Computing, Communications, and Applications}
\bvolume{18}(\bissue{1}),
\bfpage{1}--\blpage{19}
(\byear{2022})
\end{barticle}
\endbibitem

\bibitem[\protect\citeauthoryear{Jiang et~al.}{2015}]{jiang2015salicon}
\begin{bchapter}
\bauthor{\bsnm{Jiang}, \binits{M.}},
\bauthor{\bsnm{Huang}, \binits{S.}},
\bauthor{\bsnm{Duan}, \binits{J.}},
\bauthor{\bsnm{Zhao}, \binits{Q.}}:
\bctitle{Salicon: Saliency in context}.
In: \bbtitle{Proceedings of the IEEE Conference on Computer Vision and Pattern Recognition},
pp. \bfpage{1072}--\blpage{1080}
(\byear{2015})
\end{bchapter}
\endbibitem

\bibitem[\protect\citeauthoryear{Borji and Itti}{2015}]{borji2015cat2000}
\begin{botherref}
\oauthor{\bsnm{Borji}, \binits{A.}},
\oauthor{\bsnm{Itti}, \binits{L.}}:
Cat2000: A large scale fixation dataset for boosting saliency research.
arXiv preprint arXiv:1505.03581
(2015)
\end{botherref}
\endbibitem

\bibitem[\protect\citeauthoryear{Judd et~al.}{2009}]{judd2009learning}
\begin{bchapter}
\bauthor{\bsnm{Judd}, \binits{T.}},
\bauthor{\bsnm{Ehinger}, \binits{K.}},
\bauthor{\bsnm{Durand}, \binits{F.}},
\bauthor{\bsnm{Torralba}, \binits{A.}}:
\bctitle{Learning to predict where humans look}.
In: \bbtitle{2009 IEEE 12th International Conference on Computer Vision},
pp. \bfpage{2106}--\blpage{2113}
(\byear{2009}).
\bcomment{IEEE}
\end{bchapter}
\endbibitem

\bibitem[\protect\citeauthoryear{Judd et~al.}{2012}]{judd2012benchmark}
\begin{botherref}
\oauthor{\bsnm{Judd}, \binits{T.}},
\oauthor{\bsnm{Durand}, \binits{F.}},
\oauthor{\bsnm{Torralba}, \binits{A.}}:
A benchmark of computational models of saliency to predict human fixations
(2012)
\end{botherref}
\endbibitem

\bibitem[\protect\citeauthoryear{Lautenbacher}{2012}]{lautenbacher2012still}
\begin{botherref}
\oauthor{\bsnm{Lautenbacher}, \binits{O.P.}}:
From still pictures to moving pictures.
Eye-tracking in Audiovisual Translation,
135--155
(2012)
\end{botherref}
\endbibitem

\bibitem[\protect\citeauthoryear{Cerf et~al.}{2007}]{cerf2007predicting}
\begin{botherref}
\oauthor{\bsnm{Cerf}, \binits{M.}},
\oauthor{\bsnm{Harel}, \binits{J.}},
\oauthor{\bsnm{Einh{\"a}user}, \binits{W.}},
\oauthor{\bsnm{Koch}, \binits{C.}}:
Predicting human gaze using low-level saliency combined with face detection.
Advances in neural information processing systems
\textbf{20}
(2007)
\end{botherref}
\endbibitem

\bibitem[\protect\citeauthoryear{Yu et~al.}{2010}]{yu2010comparing}
\begin{barticle}
\bauthor{\bsnm{Yu}, \binits{D.}},
\bauthor{\bsnm{Park}, \binits{H.}},
\bauthor{\bsnm{Gerold}, \binits{D.}},
\bauthor{\bsnm{Legge}, \binits{G.E.}}:
\batitle{Comparing reading speed for horizontal and vertical english text}.
\bjtitle{Journal of vision}
\bvolume{10}(\bissue{2}),
\bfpage{21}--\blpage{21}
(\byear{2010})
\end{barticle}
\endbibitem

\bibitem[\protect\citeauthoryear{Singh}{2006}]{singh2006impact}
\begin{barticle}
\bauthor{\bsnm{Singh}, \binits{S.}}:
\batitle{Impact of color on marketing}.
\bjtitle{Management decision}
\bvolume{44}(\bissue{6}),
\bfpage{783}--\blpage{789}
(\byear{2006})
\end{barticle}
\endbibitem

\end{thebibliography}

\end{document}